\RequirePackage[hyphens]{url}

\documentclass[twocolumn]{svjour3}          
\smartqed  
\usepackage{graphicx}
\usepackage{cite}

\usepackage[breaklinks=true]{hyperref}
\usepackage{breakcites}
\usepackage{bm,amsmath,amssymb} 
\usepackage{psfrag}
\usepackage{natbib}
\usepackage{color,soul} 

\journalname{Author's version, published in IJCV}

\newcommand{\Figs}{Figs/}

\newcommand{\ie}{\mbox{\emph{i.e.\ }}}
\newcommand{\wrt}{\mbox{\emph{w.r.t.\ }}}
\newcommand{\eg}{\mbox{\emph{e.g.\ }}}

\newcommand{\Eq}[1]{Eq.\,(\ref{#1})}  
\newcommand{\Fig}[1]{Fig.\,\ref{#1}}  
\newcommand{\Sec}[1]{Sec.\,\ref{#1}}  
\newcommand{\Tab}[1]{Tab.\,\ref{#1}}  

\newcommand{\M}[1]{\mathtt{#1}}   
\newcommand{\V}[1]{\mathbf{#1}}   
\newcommand{\C}[1]{\mathcal{#1}}  

\newcommand{\MM}[1]{\textsc{#1}}  

\newcommand{\imu}{\textsc{imu}}  

\begin{document}

\title{Renormalization for Initialization of Rolling Shutter Visual-Inertial Odometry}



\author{Branislav Micusik        \and
            Georgios Evangelidis 
}


\institute{Snap Inc. \at
             Fleischmarkt 3-5, 1010 Vienna, Austria, 
             \email{brano@snap.com, georgios@snap.com}          
}

\date{}

\maketitle

\begin{abstract}
	In this paper we deal with the initialization problem of a visual-inertial odometry system with rolling shutter cameras. Initialization is a prerequisite for using inertial signals and fusing them with visual data. We propose a novel statistical solution to the initialization problem on visual and inertial data simultaneously, by casting it into the renormalization scheme of Kanatani. The renormalization is an optimization scheme which intends to reduce the inherent statistical bias of common linear systems. We derive and present the necessary steps and methodology specific to the initialization problem. Extensive evaluations on ground truth exhibit superior performance  and a gain in accuracy of up to $20\%$ over the originally proposed Least Squares solution. The renormalization performs similarly to the optimal Maximum Likelihood estimate,  despite arriving at the solution by different means. With this paper we are adding to the set of Computer Vision problems which can be cast into the renormalization scheme.
 

\end{abstract}

\section{Introduction}

\begin{figure}[t]
  \begin{center}
    \psfrag{i}[bl][bl]{$\tau_i$}
    \psfrag{j}[tl][bl]{$\tau_j$}
    \psfrag{0}[br][bl]{$\C{O}$}
    \psfrag{v0}[Bl][bl]{$\V{v}_0$}
    \psfrag{g}[br][br]{$\V{g}_0$}
    \psfrag{IMU}[bl][bl][1][-15]{\imu{}}
    \includegraphics[width=0.95\linewidth]{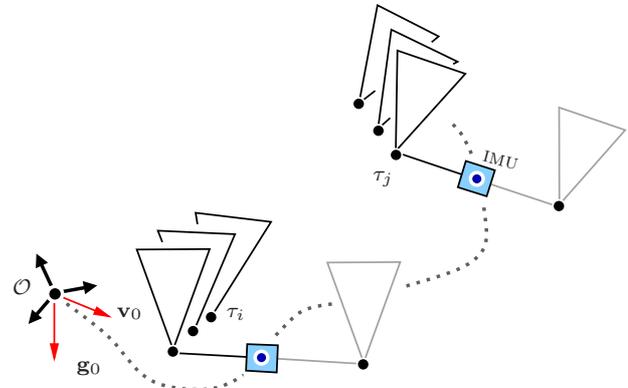} \\\vspace*{-0.5cm}
  \end{center}
 \caption{Rolling shutter (stereo) camera with \imu{} rig in motion. This paper deals with the simultaneous estimation of the unknown initial velocity $\V{v}_0$ and the gravity $\V{g}_0$ from the \imu{} and camera data streams.}
 \label{fig:motivation}
\end{figure}

Real time pose estimation of a moving camera has been an active topic in Computer Vision and Robotics community for decades, but with the advent of Augmented Reality, the topic experiences a new hype. Augmented Reality draws new requirements on pose estimation performance as on the energy consumption side as well as on the accuracy and robustness. Mobile phones, wearable smart glasses or watches use in-built rigs with a mono or stereo camera and an \imu{} as a de facto standard hardware. This stems from the fact that combining the two sensor modalities, the visual and inertial one, has been proven to be an ultimate solution towards compensating each others drawbacks. Common development toolkits natively support fusion of both sensors, \eg ARCore (\cite{ARCore}) and ARKit (\cite{ARKit}). 

There are two important practical challenges to be considered. First, cameras in mobile devices are in majority rolling shutter cameras. These are cheaper and possess higher dynamic range than standard global shutter cameras. Majority of research has been, however, devoted to the standard global shutter camera models. Rolling shutter geometry has started catching the attention with the rise of mobile phones and smart glasses (\cite{Meingast-CoRR2005,Hedborg-CVPR2012,Albl-CVPR2015,Dai-CVPR2016,Albl-CVPR2016}). Second, \imu{} sensors do measure linear acceleration and angular velocity which are the second and the first derivatives of the desired position and orientation, respectively. In order to estimate the position of an \imu{} over time, the integration of both signals needs to be performed. Such an integration requires knowing the initial conditions, \ie the initial velocity and gravity direction. However, the \imu{} data alone is insufficient to estimate these initial conditions. For instance, the \imu{} delivers zero inertial acceleration signal when it is either static or moving with constant velocity. Without proper initial conditions, the integration leads to the same static pose. The remedy lies in fusion of the inertial with visual data as the visual cues clearly distinguish these two cases.

Yet, most systems assume that the mobile device is static at the beginning of its operation and as such, the initial velocity can be set to zero and the initial gravity direction can be deduced from the accelerometer. While this assumption may be safe in many situations, it is violated when triggering the start of a Visual-Inertial Odometry (VIO) system under motion, \eg when walking or bicycling.

\subsection{Contribution}
In this paper we aim at estimating the initial velocity and gravity direction of a moving rig, equipped with rolling shutter cameras and an \imu{}, as depicted in \Fig {fig:motivation}. We consolidate the closed-form minimal solver of \cite{Martinelli-IJCV2013} for a general case with multiple rolling shutter cameras and partial tracks. As the main contribution, we introduce a more accurate solution by casting the original formulation into the renormalization scheme of \cite{Kanatani-Stat1996}. Namely, we reduce the problem of the original solver by Shur complement based elimination and present noise propagation analysis on the reduced problem in order to arrive at the renormalization scheme. 

The renormalization scheme of Kanatani is a statistical method for certain type of problems. We show that the initialization problem can be brought by the proposed operations into the renormalization scheme. The proposed solution has superior performance to the least squares solver of \cite{Martinelli-IJCV2013} while both defined on top of the same linear system. The renormalization scheme performs comparably and sometimes outperforms the optimal Maximum Likelihood (ML) estimator which minimizes the re-projection error. Compared to the least squares, the renormalization scheme removes its inherent bias, and explicitly provides the covariance of the estimate. The renormalization scheme may suffice to solve the problem in most cases, however, it can initialize ML to enforce faster convergence. 

It is known that ML entails statistical bias in the presence of what is known as “nuisance parameters”. Various studies exist for analyzing and removing bias in the ML solution, \eg by \cite{Okatani-CVPR2009}. An optimal ML solution is usually given by a nonlinear optimization which is time-consuming when solved by numerical search. It often requires extra nuisance parameters, initial values and solving iteratively large, although sparse, linear systems. On the contrary, the renormalization procedure requires neither a priori knowledge of the initial values nor the noise level which is estimated a posteriori as a result of the renormalization itself. It consists of iterated computations of eigenvalues and eigenvectors of small matrices and bias-correction steps.

This paper adds a new problem into the set of Computer Vision problems which can be cast into the renormalization scheme. \cite{Kanatani-Guide3D} formalized the renormalization scheme for many geometric computations in computer vision, \eg ellipse, homography, fundamental matrix fitting, triangulation, and 3D reconstruction. These techniques show superior performance under some circumstances on many problems to the Gold Standard Methods of \cite{Hartley-Book2004} and are viable alternatives in many practical use cases. 

The paper is structured as follows. First, the related work for VIO initialization and structure from motion systems as well as positioning the renormalization scheme is reviewed in \Sec{sec:related_work}.  Then, the main concept is presented in \Sec{sec:concept}. Its main parts include geometric relation of a camera and an \imu{} in \Sec{sec:geometry}; rolling shutter image formation in \Sec{sec:rolling_shutter}; the closed-form minimal and overconstrained solver, adjusted for rolling shutter cameras and partial feature tracks in \Sec{sec:min_solver}; its reduced form in \Sec{sec:reduced_min_solver}, cast as the renormalization scheme outlined and applied in \Sec{sec:renorm}. Bundle adjustment is shortly discussed and compared to the renormalization in \Sec{sec:renorm_vs_ba}. Finally, an extensive experimental evaluation is presented in \Sec{sec:experiments}, followed by the conclusion in \Sec{sec:conclusion}.
 
\section{Related Work}
\label{sec:related_work}

\subsection{VIO Initialization}
The most relevant paper to our method is the closed-form solution for initial velocity and gravity direction by \cite{Martinelli-IJCV2013}. The method assumes a mono global shutter camera and complete tracks. They propose to relate corresponding visual observations through the camera baseline, which linearly depends on the unknown state parameters, that is, the velocity, the gravity in the \imu{} frame and the accelerometer bias. Each visual correspondence contributes three linear equations, while the distances between map points and the cameras become unknown parameters too. The resulting linear system is solved with the constraint on the gravity magnitude. The robustness of the method against biased \imu{} readings was investigated by~\cite{Kaiser-RAL2017} and, to account for the gyroscope bias, a non-linear refinement method was proposed. \cite{Campos-ICRA2019} then built on \cite{Martinelli-IJCV2013} and \cite{Kaiser-RAL2017}, and improved the method via multiple loops of visual-inertial bundle adjustments and consensus tests. Our solution improves the least square solution and could be directly used for the bundle adjustment initialization of \cite{Campos-ICRA2019} or \cite{Mur-Artal-RAL2017}. The proposed methodology could also be applied to the reduced linear solver for initial velocity and gravity direction by \cite{Evangelidis-RAL2021}.


The above methods adopt an early fusion approach, \ie a tightly-coupled fusion. Instead, the visual SfM problem can be first solved and the \imu{} data can be later integrated in a more loosely-coupled framework of \cite{Kneip-IROS2011,Mur-Artal-RAL2017,Huang-TRO2020}. In this context, \cite{Kneip-IROS2011} suggested using visual SfM to obtain camera velocity differences which are then combined with integrated \imu{} data to recover the scale and gravity direction. The initialization part of \cite{Mur-Artal-RAL2017} used scaleless poses from ORB-SLAM of \cite{Mur-Artal-TRO2015} and then solved several sub-problems to initialize the state and the biases along with the absolute scale. This multi-step solution for the parameter initialization was then adapted in~\cite{Qin-IROS2011}.

The initialization problem becomes harder when the device is uncalibrated (\cite{Dong-IROS2012,Huang-TRO2020}). Even if the biases are known or ignored, the unknown orientation between the camera and the \imu{} makes the model non-linear and iterative optimization is necessary. \cite{Dong-IROS2012} propose two solutions to estimate the unknown orientation, thus allowing solving a linear system which, in turn, initializes a non-linear estimator. Instead, \cite{Huang-TRO2020} builds on the mutli-step approach of \cite{Mur-Artal-RAL2017} to jointly calibrate the extrinsics and initialize the state parameters. In a real scenario, however, the joint solution of calibration and initialization problem using only the very first few frames might make the pose tracking algorithm prone to diverge. 

It is worth noting that all the above works assume that visual observations come from a global-shutter sensor. Consumer devices, however, are mostly equipped with rolling shutter cameras and rolling-shutter effects need to be handled. Proper treatment of the rolling shutter camera in connection to visual-inertial odometry can be found in work of  \cite{Hedborg-CVPR2012,Li-ICRA2013,Patron-Perez-IJCV2015,Bapat-CVPR2018,Ling-ECCV2018,Schubert-ECCV2018,Schubert-IROS2019}. However, neither of the works copes with the initialization problem. 

\subsection{Renormalization}
The renormalization of \cite{Kanatani-Stat1996} was at first not well accepted by the computer vision community. This was due to the generally held preconception that parameter estimation should minimize some cost function. Scientists wondered what renormalization was minimizing. In this line of thought, \cite{Chojnacki-JMIV2001} interpreted renormalization as an approximation to ML. Optimal estimation does not necessarily imply minimizing a cost function and as such the renormalization is an effort to improve accuracy by a direct mean (\cite{Kanatani-ICPR2014}). The mathematical foundation of the optimal correction techniques of \cite{Kanatani-Guide3D} is also discussed in the broader scope of photogrammetric statistical geometric computations by \cite{Foerstner-Book2016}. It is the non-minimization formalism based on error analysis which intuitive meaning is often difficult to grasp, as we will see in the following.

Regarding re-projection error minimization as the ultimate method, or the Gold Standard, the fact that the accuracy of ML can be improved was rather surprising~(\cite{Kanatani-IJCV2008,Okatani-CVPR2009}). For hyperaccurate correction, however, one first needs to obtain the ML solution by an iterative method such as Fundamental Numerical Scheme (FNS) of \cite{Chojnacki-PAMI2000} on Sampson Error or Heteroscedastic Error-In-Variables (HEIV)  method of \cite{Leedan-IJCV2000} and also estimate the noise level. However, it is possible to directly compute the corrected solution from the beginning, by modifying the FNS iterations if one adopts the non-minimization approach of geometric estimation of \cite{Kanatani-ICPR2014}.

\section{Concept}
\label{sec:concept}

\begin{figure}[t]
  \begin{center}
    \psfrag{i}[bl][bl]{$\tau_i$}
    \psfrag{j}[bl][bl]{$\tau_j$}
    \psfrag{pi}[bl][bl]{$\lambda_i \V{p}_i$}
    \psfrag{pj}[Br][Bl]{$\lambda_j \V{p}_j$}
    \psfrag{imu}[tl][bl]{\imu{} samples}
    \psfrag{0}[Bc][Bl]{$\C{O}$}
    \psfrag{ti0}[br][bl]{$\V{t}_i^0$}
    \psfrag{tj0}[bl][bl]{$\V{t}_j^0$}
    \psfrag{tcj}[bl][bl]{$\M{R}_j^0 \V{t}_\MM{c}^\imu{}$}
    \psfrag{err}[bl][bl]{$\epsilon$}
  \includegraphics[width=.6\linewidth]{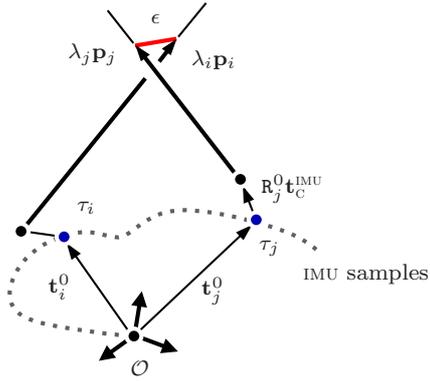} 
  \end{center}
 \caption{Distance error $\epsilon$ being minimized by the na{\"i}ve least square estimator.}
 \label{fig:distance_err}
\end{figure}

\subsection{Geometry}
\label{sec:geometry} 

A 3D point $\V{X}_i$ expressed in the local coordinate system of the \imu{} at time $\tau_i$, projected into the coordinate system of the \imu{} at time $\tau_0$ reads as
\begin{align} 
\V{X}_0 &= \M{R}_{i}^0 \, \V{X}_i + \V{t}_i^0,
\end{align}
where $\M{R}_{i}^0$ and $\V{t}_i^0$ stand for the rotation matrix and the translation vector to perform this transformation. Let us assume that a camera attached to the \imu{} rig observes the 3D point $\V{X}_0$ at time $\tau_i$ as 
\begin{align}
  \lambda_i \V{u}_i = \M{K} \, ( \M{R}^\MM{c}_\imu{} \, \M{R}_0^i \, &\V{X}_0 + \M{R}^\MM{c}_\imu{} \V{t}_0^i + \V{t}^\MM{c}_\imu{}) \notag\\
  \lambda_i \underbrace{\M{R}_i^0 \M{R}_\MM{c}^\imu{} \M{K}^{-1} \V{u}_i}_{\tilde{\V{p}}_i} + \V{t}_i^0 + \M{R}_i^0 \V{t}_\MM{c}^\imu{}  = \, &\V{X}_0 %
  \label{eq:triangulation}
\end{align}
where $\M{R}_\MM{c}^\imu{}$, $\V{t}_\MM{c}^\imu{}$ is the known fixed relative pose from the camera to the \imu{}, and $\M{K}$ is the known camera calibration matrix. Image coordinates of the 3D point $\V{X}_0$, being tracked in multiple views, are denoted $\V{u}_i$, and $\lambda_i$ are unknown scales, the depths, of their projection rays $\M{K}^{-1} \V{u}_i$.

The \imu{} pose $\M{R}_i^0$, $\V{t}_i^0$ at time $\tau_i$ is calculated as
\begin{align}
  \M{R}_i^0 &= \prod_{k=0}^{i-1} \M{R}_{k+1}^{k} = \prod_{k=0}^{i-1} \M{\Omega}(\bm{\omega}_{k} \, \Delta\tau), \label{eq:imu_rot} \\
  \V{t}_i^0 &= \V{t}_0 + i \, \V{v}_0 \Delta \tau + \left( \sum_{k=0}^{i-1} \beta_{k,i} \, \M{R}_k^0 \, \V{a}_k + i^2 \V{g}_0 \right) \frac{{\Delta\tau}^2}{2},
  \label{eq:imu}
\end{align}
where
\begin{equation}
\beta_{k,i} = 2i - 2k - 1.
\end{equation}
The $3$ element vector $\V{a}_k$ and $\bm{\omega}_k$ is the accelerometer and the gyroscope readout measurements of the \imu{} at time $\tau_k$, respectively. The exponential map $\M{\Omega}(.)$ gives a rotation matrix from the argument vector. The time between two \imu{} samples is denoted by $\Delta \tau$. Without loss of generality, we set the origin  into the coordinate system of the first \imu{}, thus the translation $\V{t}_0 = \V{0}$. The initial velocity $\V{v}_0$ and the gravity vector $\V{g}_0$ at time $\tau_0$, expressed in the origin, are the unknowns and subjects to estimation. For the sake of simplicity, we assume for now that the measurements are corrected for biases. The compensation of the biases is discussed later in \Sec{sec:biases}. The biases may vary over time, and can be included in a final non-linear refinement step. We further assume that the \imu{} and the cameras are temporarily synchronized.

It is to be noted that unless $\V{v}_0$ and $\V{g}_0$ are known, the \imu{} data cannot be integrated in order to get \imu{} poses in the above chosen origin. Most visual-inertial systems assume that the camera-\imu{} rig is {\em static} at start and it can be assumed that the initial velocity $\V{v}_0 = \V{0}$ and the initial gravity $\V{g}_0$ is determined from the acceleration readout. However, in many practical situations this is violated, the system is in motion at start, \eg a person rides a bicycle or walks.

\subsection{Rolling shutter image formation}
\label{sec:rolling_shutter}

A rolling shutter camera is, in its principle, a moving line camera. When moving along a line, it falls into a class of linear pushbroom cameras, see~(\cite{Gupta-PAMI1997}). Each scanline is read out one after the other and all of them are stacked into an image buffer. Note that indeed the pose of the \imu{} in \Eq{eq:triangulation} differs for each $i$. The readout time of a scanline of the rolling shutter camera is constant even when camera exposure varies. We can therefore safely choose $\Delta \tau = \tau_{i+1} - \tau_i$ to be exactly the readout time of one line of the camera. The \imu{} data can be upsampled, \eg for VGA resolution from a typical sampling \imu{} rate of 800Hz to 47.6kHz, and integrated, called the {\em interpolate-then-integrate} approach. As such, for each scanline of the image, we have one pose,  $\M{R}_i^0$ and $\V{t}_i^0$. Alternatively, the integration is performed on the original \imu{} sample rate and then the poses are interpolated for each scanline, the {\em integrate-then-interpolate} approach. We found the first approach to give slightly better results for the initialization problem. This is expected because of the non-linear dependency of translation $\V{t}_i^0$ on gyroscope readout $\bm{\omega}_k$ in \Eq{eq:imu} and \Eq{eq:imu_rot}. Upsampling the signals first and then integrating through a non-linearity is typically recommended.





\subsection{Linear Solver}
\label{sec:min_solver}

Let us assume that a (stereo) camera with the \imu{} moves and observations of some 3D points in multiple images are available. If $\V{u}_i$ and $\V{u}_j$ are the homogeneous image observations of a point $\V{X}_0$ in two views, then we can write \Eq{eq:triangulation} for each point separately. By eliminating $\V{X}_0$ we obtain
\begin{equation}
 \lambda_i \V{p}_i + \V{t}_i^0 + \M{R}_i^0 \V{t}_\MM{c}^\imu{} = \lambda_j \V{p}_j + \V{t}_j^0 + \M{R}_j^0 \V{t}_\MM{c}^\imu{},
  \label{eq:two_view_constraint}
\end{equation}
such that the $3$ element calibrated vector $\V{p}_i = \C{N}(\tilde{\V{p}}_i) = \C{N}(\M{R}_i^0 \M{R}_\MM{c}^\imu{} \M{K}^{-1} \V{u}_i)$, where $\C{N}(\V{x})$ normalizes the vector $\V{x}$ by its third coordinate to the homogeneous coordinates.
Substituting \Eq{eq:imu} into \Eq{eq:two_view_constraint} yields

{\small
\begin{align}
\arraycolsep=5pt
\left[\begin{array}{c|c}
\underbrace{%
\begin{matrix}
 \xi_{ij} \M{I}_3 & \mu_{ij} \M{I}_3 & \bm{\kappa}_{ij}\\
 \vdots \\
 \xi_{ik} \M{I}_3 & \mu_{ik} \M{I}_3 & \bm{\kappa}_{ik}  \\
 \vdots \\
 \xi_{kl} \M{I}_3 & \mu_{kl} \M{I}_3 & \bm{\kappa}_{kl}
\end{matrix}}_{\displaystyle \M{S}} &
\underbrace{%
\begin{matrix}
   \V{p}_{i} & \V{p}_{j} & \bm{.} & \ldots & \bm{.} \\
   \vdots \\
   \V{p}_{i} & \bm{.} & \V{p}_{k} & \ldots & \bm{.} \\
   \vdots \\
  \bm{.} & \bm{.} & \V{p}_{k} & \ldots & \V{p}_{l}  
\end{matrix}}_{\displaystyle \M{P}} 
\end{array}\right] \left[\begin{array}{c}\V{v}_0 \\ \V{g}_0 \\ 1 \\ \lambda_i \\ \lambda_j \\ \lambda_k \\ \vdots \\ \lambda_l \end{array}\right] & = \V{0},
\label{eq:lin_solver}
\end{align}}
where 
\begin{align}
\xi_{ij} & = (i - j) \Delta \tau, \notag\\
\mu_{ij} & = \left(i^2 - j^2\right) \frac{{\Delta\tau}^2 }{2}, \notag\\
\bm{\kappa}_{ij} & = \M{R}_i^0 \V{t}_\MM{c}^i - \M{R}_j^0 \V{t}_\MM{c}^j + \notag\\
&\hspace*{8mm} + \left( \sum_{k=0}^{i-1} \beta_{k,i} \, \M{R}_k^0 \, \V{a}_k - \sum_{k=0}^{j-1} \beta_{k,j} \, \M{R}_k^0 \, \V{a}_k \right)\frac{{\Delta\tau}^2 }{2}.
\end{align}
In the matrix form, the \Eq{eq:lin_solver} can be written as
\begin{equation}
[\M{S} \ \ \M{P}] \, \V{x} = \V{0},
\label{eq:lin_solver_matrix_form}
\end{equation}
which is a linear equation system. It can be solved, for instance, in the least squares sense. It is worth noting that the error which is minimized by the above least squares solution has a geometric meaning. It relates to the distance between 3D points which are obtained through $\lambda_i \V{p}_i$, as shown in~\Fig{fig:distance_err}. We tried to formulate the initialization problem on the angular error on projective rays instead of the distance. The angular error is often used in standard epipolar geometry solvers (\cite{Hartley-Book2004}), and also has been used in the relative pose for the rolling shutter camera in~\cite{Dai-CVPR2016}. For static or slow motion the error degenerates as is too sensitive to image noise. Overall, the angular error is inferior to the presented distance based error.

\begin{figure}[t]
\begin{center}
  \rotatebox{90}{\quad \quad \# of samples} 
  \psfrag{pixel shift}[c][c]{\footnotesize{pixel shift}}
  \includegraphics[width=0.5\linewidth, trim = 1 0 0 1, clip]{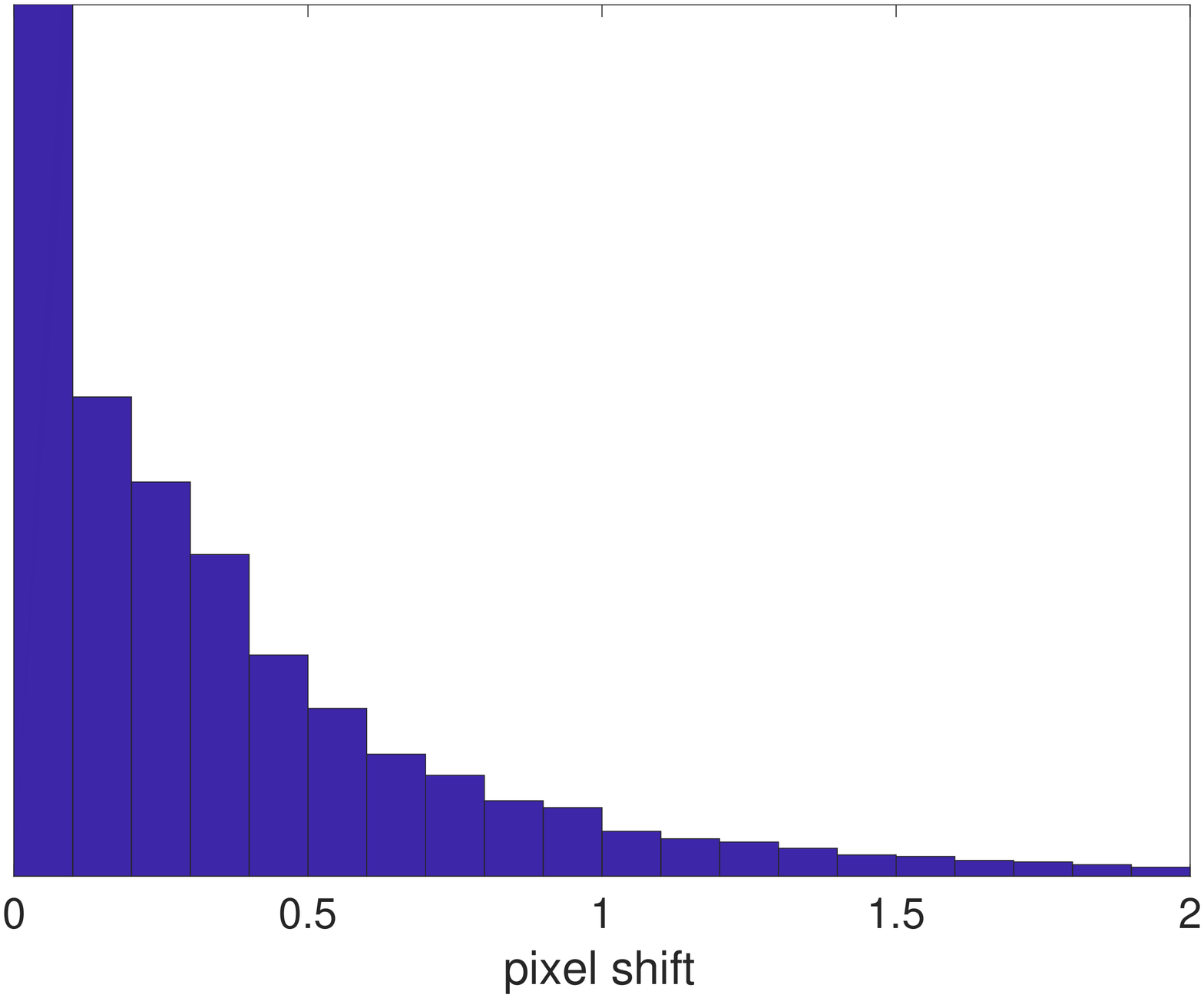} 
\end{center}
\caption{Half-normal distribution of the positional error of detected FAST corners of \cite{Rosten-PAMI2008} and ECC tracked feature points by \cite{Evangelidis-PAMI2008}. The error is defined as the distance between a feature point and its ground truth location on a synthetically rendered sequence with no additionally added image noise. It can be seen that it obeys Gaussian distribution under a perfect image formation model. For real images we silently assume similar behavior.}
\label{fig:gaussian_noise}
\end{figure}

Each matching pair of image points adds three equations which constrain the shared unknown initial velocity and gravity (fixed six unknowns), but adds additional unknown $\lambda$'s per ray (always two new unknowns). Provided that a single point is tracked in all views, then each new image observation adds only one $\lambda_i$, and the minimum number of frames is $5$ ($4$ pairs). The unknown $\lambda$'s are shared between multiple views as shown in \Eq{eq:lin_solver}, if a ray is used in multiple pairs. There, for the $(i,j)$ and $(i,k)$ pairs, $\lambda_i$ is shared as the corresponding 3D point is projected into three views. This explicit sharing of $\lambda$'s better constraints the system and reduces the growth of unknowns. Similar derivations to~\Eq{eq:lin_solver} for the global shutter camera can be found in \cite{Martinelli-IJCV2013}.


\subsection{Reduced Linear Solver}
\label{sec:reduced_min_solver}

We propose to eliminate the unknown $\lambda$'s form \Eq{eq:lin_solver_matrix_form}. This can be done with the Schur complement based elimination of the $\M{P}$ matrix in \Eq{eq:lin_solver_matrix_form} such that it becomes
\begin{align}
[ \M{S} - \underbrace{\M{P} (\M{P}^\top \M{P})^{-1} \M{P}^\top}_{\M{G}} \M{S} ] \ \V{y} &= \V{0}, \label{eq:lambda_elimination}\\
[ (\M{I}_3 - \M{G}) \, \M{S} ] \ \V{y} &= \V{0}, \label{eq:reduce_lin_solver1}\\
\M{B} \ \left[ \begin{array}{c} \V{v}_0 \\ \V{g}_0 \\ 1\end{array} \right]  &= \V{0},
\label{eq:reduce_lin_solver}
\end{align}
where the matrix $\M{B}=(\M{I}_3 - \M{G})\, \M{S}$ is ${3 N \times 7}$, $N$ is the number of pairs of point matches and $\V{y} = [\V{v}_0^\top \ \V{g}_0^\top \ 1]^\top$ is the unknown $7$ element vector. The matrix $\M{G}$ is an idempotent $N \times N$ projection matrix. Solving 
the reduced linear problem in~\Eq{eq:reduce_lin_solver} in the least square sense, yields  the same result as \Eq{eq:lin_solver}. Depending on the sparsity of the matrix $\M{P}$, one or another can be faster, and should be chosen accordingly for specific practical conditions. 

The reduced form simplifies the noise analysis in order to arrive to the solver presented in the next. The advantage of the reduced form in \Eq{eq:reduce_lin_solver} is that the unknown solution vector is of {\em fixed} size. It contains the initial velocity and gravity only, and no longer the depth multipliers $\lambda's$. 

At the first look, it might look as if the rolling shutter camera adds difficulties in the equations in comparison to the global shutter case. However, from the geometric point of view, the opposite can be claimed. Each camera line has a different center of projection when moving and so the rays put into the triangulation equation constrain better the solution. On the other side, the rolling shutter effect adds image artifacts as the long line segments may get projected as bent under a fast motion. However, as we experimentally observed, this is negligible for a feature tracker which uses a small image patch. Overall, the rolling shutter is a beneficial feature which implicitly encodes motion of the camera and makes it directly observable~(\cite{Bapat-CVPR2018}).

Both \Eq{eq:lin_solver} and \Eq{eq:reduce_lin_solver} treat all the data as equally valuable and fall into the class of algebraic least squares estimators. In the next section we propose an improvement by involving a proper noise perturbation analysis to individually weight the point matches.


\subsection{Renormalization Scheme}
\label{sec:renorm}
Class of problems like  \Eq{eq:reduce_lin_solver}, where a geometric relationship in high dimensions, expressed as an implicit equation, is fitted, is called {\em geometric fitting} and has been studied by \cite{Kanatani-Stat1996}. The matrix $\M{B}$ in \Eq{eq:reduce_lin_solver} is filled from the \imu{} sensor data and from the image point correspondences over multiple views $\V{u}_{\{i,j,\ldots\}}$ via their projective rays $\V{p}_{\{i,j,\ldots\}}$. We consider short integration time in which the effect of noise on the \imu{} data is negligible to the noise on the point correspondences. We experimentally verified very little accuracy gain when considering noise in the \imu{} data. Therefore, in the next we perform noise perturbation analysis when considering noise purely on the point correspondences.

Each image point correspondence pair $\V{u}_i \leftrightarrow\V{u}_j$ in \Eq{eq:reduce_lin_solver} contributes three equations to the matrix $\M{B}$ and can be written as
\begin{equation*}
\big( \V{b}^{(s)}(\V{u}_i, \V{u}_j), \V{y} \big) = 0, \quad \quad s=1 \ldots 3,
\end{equation*}
where $\V{b}^{(s)}(.,.)$ is one row of the matrix $\M{B}$ and $(\V{a}, \V{b}) = \V{a}^\top\,\V{b}$ stands for the inner product. The three equations are linearly dependent, so we see the same as in \Eq{eq:lin_solver_matrix_form} that $6$ is the minimal number of point correspondences to guarantee $rank(\M{B})=6$. 

The coordinates of the image point correspondences $\V{u}_i, \V{u}_j$ are not perfect. This is caused by the image operations, specifically the feature detection and patch based tracking on noisy image data signal. We model this uncertainty in statistical means. We assume that the observed image point $\V{u}_i$ stems from perturbation of the true value $\bar{\V{u}}_i$  by independent random Gaussian variable $\Delta \V{u}_i$ of zero-mean and with the covariance matrix $\M{V}[\V{u}_i]$, such that
\begin{equation*}
\V{u}_i = \bar{\V{u}}_i + \Delta \V{u}_i.
\end{equation*}
We experimentally validated that Gaussian noise is a feasible assumption in practical situations with an off-the-shelf feature detector and a feature tracker, see \Fig{fig:gaussian_noise} for more details. We assume a $2\times2$ covariance matrix, known up to noise level $\sigma$, 
\begin{equation}
\M{V}[\V{u}_i] = \sigma^2 \M{V}_0 [\V{u}_i],
\label{eq:noise_level}
\end{equation}
where the known normalized covariance matrix $\M{V}_0 [\V{u}_i]$ describes the orientation dependence of uncertainty in relative terms. The covariance matrix can come from uncertainty of the employed feature detector and the tracker. In all our experiments we assume $\M{V}_0 [\V{u}_i]$ be the identity matrix.

If the observations $\V{u}_i$, $\V{u}_j$ are regarded as random variables, their nonlinear mapping $\V{b}^{(s)}(\V{u}_i, \V{u}_j)$, which we write $\V{b}^{(s)}_{ij}$, or  $\V{b}^{(s)}_{\alpha}$ for short, is also a random variable. The linear index $\alpha$ steps over the row triplets in the matrix $\M{B}$ in \Eq{eq:reduce_lin_solver}. Each $\alpha$ represents a frame pair $ij$ and we use them exchangeably. Missing superscript in $\V{b}_{ij}$ means all three rows, \ie a $3\times7$ matrix. Its covariance matrix is
\begin{equation}
\M{V}^{(st)}[\V{b}_{ij}] = \sigma^2 \M{V}^{(st)}_0[\V{b}_{ij}],
\end{equation}
where $s, t =1\ldots 3$ to combine mutually the rows of the three equations per correspondence, yielding nine matrices per each $ij$ combination (or $\alpha$). The covariance matrix is evaluated to first approximation in terms of the Jacobians $\M{J}^{(s)}$ and $\M{J}^{(t)}$ of the mapping $\V{b}_{ij}$ as follows
\begin{equation}
\M{V}^{(s t)}_0[\V{b}_{ij}]_{_{7\times7}} = \M{J}^{(s)}_{ij} \, \left[  \begin{array}{cc} \sigma^2 \M{V}_0 [\V{u}_i] & \M{0} \\ \M{0} & \sigma^2 \M{V}_0 [\V{u}_j] \end{array} \right]_{4\times4}  \M{J}_{ij}^{(t)^\top}.
\label{eq:cov_propagation}
\end{equation}

If the noise in the $\V{u}$-space is assumed Gaussian, the corresponding noise in the transformed $\V{b}$-space is no longer Gaussian. However, our numerical experiments have shown that in the noise range of typical feature detector and tracker, \ie $\sigma\in[0, 0.5]$\,pixels,  such an assumption is feasible.  In order to stay in the safe range, removal of systematic error like outliers prior to estimation is crucial. Correction for higher order noise terms can be omitted, as we observed that the Hyper-renormalization of \cite{Kanatani-Guide3D} brings only small accuracy gain for the increased computational burden.

\subsubsection{Solver} 
\label{sec:renorm_solver}

The standard Least Squares (LS) solution to \Eq{eq:reduce_lin_solver} 
\begin{align}
\epsilon^2 &= \frac{1}{N} \sum_{\alpha=1}^{N}\sum_{s,t=1}^{3} (\V{b} _\alpha^{(s)}, \V{y}) (\V{b} _\alpha^{(t)}, \V{y})\\
&= \left(\V{y},  \big(\frac{1}{N} \sum_{\alpha=1}^{N}\sum_{s,t=1}^{3} \V{b} _\alpha^{(s)} {\V{b} _\alpha^{(t)}}^\top\big)  \, \, \V{y}\right) = (\V{y}, \M{M}_{\MM{ls}}  \V{y})
\label{eq:LS}
\end{align}
minimizes the mean square error $\epsilon^2$. \Fig{fig:distance_err} depicts the geometric meaning of the error. The solution can be obtained as an eigenvalue fit of the $7 \times 7$ matrix $\M{M}_{\MM{ls}}$ of
\begin{equation*}
\M{M}_{\MM{ls}} \, \V{y} = \lambda \V{y}.
\end{equation*}
Weighting each pair $i$, $j$ differently, LS would turn, for small accuracy gain, into iterative re-weighted LS. More importantly, both can be fairly improved by \cite{Taubin-PAMI1991}, as modification of LS and even slightly more by
\cite{Kanatani-IJCV2008}. Our experiments validate what has been demonstrated in the ellipse fitting problem by \cite{Kanatani-IJCV2008}, that the error on the estimated entities can be sorted as na{\"i}ve LS $>$ weighted LS $\gg$ Taubin $>$ renormalization, see \Sec{sec:syn_experiments} and \Fig{fig:result_syn_Taubin}. 

\cite{Taubin-PAMI1991} proposed to include higher noise error terms to remove the bias of LS, and such, to first order approximation of the algebraic mean square error it yields a generalized eigenvalue fit. Kanatani further improved upon this idea and proposed to iteratively re-weight 
the Taubin method, therefore called {\em renormalization} (\cite{Kanatani-Stat1996}). In the following we present the renormalization scheme applied to the initialization of a VIO system.

{\paragraph{\bf Renormalization Scheme}
\begin{enumerate} 
\item Let $\V{y}_0=\V{0}$ and $w_\alpha^{(s t)}=\delta_{s t}$, $\alpha=1 \ldots N$, $s, t = 1,2,3$, where $\delta_{st}$ is the Kronecker delta, equal $1$ if $s = t$ and $0$ otherwise.
\item Compute $7 \times 7$ matrices 
\begin{align}
\M{M} &= \frac{1}{N} \sum_{\alpha=1}^{N}\sum_{s,t=1}^{3} {w_\alpha^{(s t)} \V{b} _\alpha^{(s)}} {\V{b} _\alpha^{(t)}}^\top %
\label{eq:renorm_M_matrix}\\
\M{N} &= \frac{1}{N} \sum_{\alpha=1}^{N}\sum_{s,t=1}^{3} w_\alpha^{(s t)} \M{V}_0^{(s t)}[\V{b}_\alpha],
\label{eq:renorm_N_matrix}
\end{align}
where $w_\alpha^{(st)}$ is the element of the matrix $\M{W}_\alpha$ at the row $s$ and column $t$.
\item Solve the generalized eigenvalue problem
\begin{equation}
\M{M} \, \V{y} =  \gamma \M{N} \, \V{y}
\label{eq:gep}
\end{equation}
and compute the unit eigenvector $\V{y}$ for the smallest eigenvalue $\gamma$.
\item If $\V{y} \approx \V{y}_0$ up to sign, continue to Step 5. Else, update
\begin{align}
\M{W}_\alpha & \leftarrow \left[\scriptsize\begin{array}{c@{\hspace{2mm}}c@{\hspace{2mm}}c} %
(\V{y}, \M{V_0^{(1 1)}[\V{b}_\alpha]} \, \V{y}) &%
(\V{y}, \M{V_0^{(1 2)}[\V{b}_\alpha]} \, \V{y})  &%
(\V{y}, \M{V_0^{(1 3)}[\V{b}_\alpha]} \, \V{y}) \\
(\V{y}, \M{V_0^{(2 1)}[\V{b}_\alpha]} \, \V{y}) &%
(\V{y}, \M{V_0^{(2 2)}[\V{b}_\alpha]} \, \V{y})  &%
(\V{y}, \M{V_0^{(2 3)}[\V{b}_\alpha]} \, \V{y}) \\
(\V{y}, \M{V_0^{(3 1)}[\V{b}_\alpha]} \, \V{y}) &%
(\V{y}, \M{V_0^{(3 2)}[\V{b}_\alpha]} \, \V{y})  &%
(\V{y}, \M{V_0^{(3 3)}[\V{b}_\alpha]} \, \V{y}) 
\end{array}\right]_{\{1,2\}}^{-}\notag\\
\V{y}_0 & \leftarrow \V{y}
\end{align}
and go back to Step 2. The expression $[\M{U}]_{\{1,2\}}^{-}$ is the pseudoinverse with truncated rank 2 or 1. The truncation to rank $2$ is done iff  $\frac{\sigma_2}{\sigma_1}>0.1$, where $\sigma_1$ and $\sigma_2$ is the first and the second largest singular value of $\M{U}$ respectively. Otherwise, the truncation to rank $1$ is performed.
\item Return $\V{y}$ composed of $\V{v}_0$ and $\V{g}_0$, its covariance matrix $\M{V}_0[\V{y}] $, and the noise level $\sigma$
\begin{equation}
\M{V}_0[\V{y}] = \frac{\sigma^2}{N} \M{J}_H \M{M}^{-1} \M{J}_H^\top, \quad \quad \quad
\sigma^2 = \frac{\V{y}^\top \M{M} \V{y}}{2-6/N}
\label{eq:cov_noise_final}
\end{equation}
with $\M{J}_H$ being the Jacobian of the transformation from a homogeneous to Euclidean vector, see~(\cite{Foerstner-Book2016}, Eq.(10.32)), 
\begin{equation*}
\M{J}_H=\frac{1}{\V{y}_{(7)}^2}[\V{y}_{(7)} \M{I}_6 \ | \ -\V{y}_{(1:6)}].
\end{equation*} 
Justification of estimating the noise level $\sigma$ can be seen in Eq.\,(6.46) in \cite{Kanatani-Guide3D}.
\end{enumerate}
}

\noindent
The matrix $\M{M}$ determines the covariance of the final estimate of $\V{y}$, while the matrix $\M{N}$ controls the bias of $\V{y}$. The contribution of the renormalization scheme is the matrix $\M{N}$. Its combination with the matrix $\M{M}$ compensates for the statistical bias which is inherent in Least Squares solution (\cite{Kanatani-IJCV2008}).

Least Squares choose $\V{y}$ which minimizes the cost function $\epsilon^2$ in \Eq{eq:LS}. In renormalization scheme there is no explicit cost function which is minimized. The estimated $\V{y}$ is obtained by solving a set of equations in order to reduce the dominant bias of optimally weighted Least Squares, such that it reaches Kanatani-Cramer-Rao lower bound (\cite{Kanatani-Stat1996}). 

The fourth step of the above algorithm deserves more attention due to switching of the pseudoinverse's truncated rank. Based on type of the problem, typically, the rank of the pseudoinverse is kept constant during the renormalization scheme. For instance, that is the case in the most similar algorithm to ours for optimal Homography estimation of \cite{Kanatani-Guide3D}. In that problem also three equations contribute to building the pseudoinverse of the weight matrix $\M{W}_\alpha$ and this matrix is naturally of rank $2$. This comes from the fact that only two equations out of three which go into this matrix are linearly independent. In our case, the situation is similar, but not that straightforward. The $3N \times 7$ matrix $\M{B}$ in \Eq{eq:reduce_lin_solver} has rank $6$ with many linearly dependent rows. Each row triplet which goes into the matrix $\M{B}$ is created through the elimination of lambdas in \Eq{eq:lambda_elimination}. How the original equations from $\M{S}$, $\M{P}$ matrices are used for eliminating $\lambda$'s depends on the structure of the matrix $\M{P}$ and noise conditions. This drops the rank, either in most cases to $2$, but occasionally to $1$. When not treating the edge case of rank $1$ this way, the weight matrix for the corresponding triplet may get very large weights and may cause divergence. The ratio of $0.1$ was achieved empirically in order to get good performance on all the tested sequences. We leave more rigorous theoretical understanding of this step for future work.

\subsubsection{Jacobians}
\label{sec:jacobians}

In order to compute the covariance matrix $\M{V}^{(s t)}_0[\V{b}_{ij}]$ in \Eq{eq:cov_propagation}, the Jacobian matrices $\M{J}^{(s)}_{ij} $ and $\M{J}^{(t)}_{ij}$ need to be computed. Each $7\times4$ Jacobian matrix is factored into four matrices
\begin{equation}
\M{J}^{(s)}_{ij} = \M{J}_{ij}^{(s)_4} \, \M{J}_{ij}^{(s)_3} \, \M{J}_{ij}^{(s)_2} \, \M{J}_{ij}^{(s)_1}.
\label{eq:total_jacobian}
\end{equation}
The first Jacobian captures the transformation of the point from homogeneous coordinates to the calibrated ray,
\begin{equation}
\M{J}_{ij}^{(s)_1} = \left[ \begin{array}{cc} \frac{1}{f_i} \M{I}_2 & \M{0} \\ \M{0} & \frac{1}{f_j} \M{I}_2 \end{array} \right]_{4\times4},
\end{equation}
where $f_i$ stands for the focal length of the camera which observes $\V{u}_i$ and $\M{I}$ for the identity matrix.

The second Jacobian captures rotation of the vector. Denoting $\tilde{\M{R}}_i = \M{R}_i^0 \M{R}_\MM{c}^\imu{}$ and $\tilde{\M{R}}_j = \M{R}_j^0 \M{R}_\MM{c}^\imu{}$, then
\begin{equation}
\M{J}_{ij}^{(s)_2} = \left[ \begin{array}{cc} \tilde{\M{R}}_{i(:,1:2)} & \M{0} \\ \M{0} & \tilde{\M{R}}_{j(:,1:2)} \end{array} \right]_{6\times4},
\end{equation}
where $\tilde{\M{R}}_{i(:,1:2)}$ is $3\times2$ matrix composed of the first two columns of the rotation matrix $\tilde{\M{R}}_{i}$.

The third Jacobian captures the transformation to homogeneous coordinates. This would not be in general needed, however, from computational point of view,  one avoids the need of derivative \wrt the $\tilde{p}_z$. Introducing this extra non-linearity is in practice not affecting the solution. Considering $\tilde{\V{p}}_i=\M{R}_i^0 \M{R}_\MM{c}^\imu{} \M{K}^{-1} \V{u}_i$ from \Eq{eq:triangulation}, then
\begin{equation}
\M{J}_{ij}^{(s)_3} = \left[ \small\begin{array}{cc} \frac{1}{\tilde{p}_{i,z}} [\M{I}_2 \ | -\V{p}_{i,(1:2)}] & \M{0} \\ %
\M{0} & \frac{1}{\tilde{p}_{j,z}} [\M{I}_2 \ | -\V{p}_{j,(1:2)}]  \end{array} \right]_{4\times6},
\label{eq:jacobian_s3}
\end{equation}
where $\tilde{\V{p}}_i = [\tilde{p}_{i,x} \ \tilde{p}_{i,y} \ \tilde{p}_{i,z}]^\top$, and $\V{p}_i = \C{N}(\tilde{\V{p}}_i) = [\V{p}_{i,(1:2)}^\top  \ 1]^\top$.

The fourth Jacobian captures the Schur complement based elimination of \Eq{eq:reduce_lin_solver},
\begin{equation}
\M{J}_{ij}^{(s)_4} = \left[ \frac{\partial \V{b}_{ij}^{(s)}}{\partial p_{i,x}} \ \ %
\frac{\partial \V{b}_{ij}^{(s)}}{\partial p_{i,y}} \ \ %
\frac{\partial \V{b}_{ij}^{(s)}}{\partial p_{j,x}} \ \ %
\frac{\partial \V{b}_{ij}^{(s)}}{\partial p_{j,y}}  \right]_{7\times 4}.
\end{equation}
The second dimension of four is due to the trick with homogeneous coordinates in \Eq{eq:jacobian_s3}, otherwise, it would be six. It brings an important saving as computing this Jacobian is computationally the most demanding part of the whole algorithm. This Jacobian requires to access the whole matrix $\M{B}$. Let us further investigate the partial derivative \wrt to the first component $p_{i,x}$
\begin{equation}
\frac{\partial \M{B}}{\partial p_{i,x}} = \frac{\partial (\M{I}_3 - \M{G})\,\M{S}}{\partial p_{i,x}} = - \frac{\partial \M{G}}{\partial p_{i,x}}\M{S},
\label{eq:derivative_B}
\end{equation}
as outcome from derivative of \Eq{eq:reduce_lin_solver1}. It is analogous for the rest three components. Recall that $\M{G} = \M{P} (\M{P}^\top \M{P})^{-1} \M{P}^\top$. For any non-singular square matrix $\M{A}$ the following holds (\cite{Golub-Matrix2013})
\begin{equation*}
\frac{\partial\M{A}^{-1}}{\partial \alpha} = -\M{A}^{-1} \, \frac{\partial \M{A}}{\partial \alpha} \, \M{A}^{-1}.
\end{equation*}
This allows to split inversion and derivative of the matrix. It can be computed only once as it is independent on the $\alpha$. Since $\M{G}$ is a regular idempotent projection matrix, then we can apply it to get
\begin{equation}
\frac{\partial \M{G}}{\partial p_{i,x}} = \frac{\partial \M{P}}{\partial p_{i,x}} \tilde{\M{P}} \M{P}^\top + %
\M{P}\tilde{\M{P}} \frac{\partial \M{P}^\top}{\partial p_{i,x}} - %
\M{P} \tilde{\M{P}} \frac{\partial (\M{P}^\top\M{P})}{\partial p_{i,x}} \tilde{\M{P}} \M{P}^\top,
\label{eq:Gderivative}
\end{equation}
where $\tilde{\M{P}} = (\M{P}^\top \M{P})^{-1} $. The factors $\tilde{\M{P}} \M{P}^\top$, $\M{P}\tilde{\M{P}}$ are computed only once for all the correspondences. The three partial derivatives are correspondence dependent as they depend on $i$ and $j$.  Since the matrix $\M{P}$ is very sparse and linear in $\V{p}$, the derivative matrices contain only few $1$'s depending how often $\V{u}_i$ appears in the correspondence pairs. Overall, using factorization in \Eq{eq:Gderivative} and sparse matrix calculus, the total Jacobian in \Eq{eq:total_jacobian} can be calculated very efficiently.

%

\subsection{Renormalization vs. Bundle Adjustment}
\label{sec:renorm_vs_ba}

In order to demonstrate performance of the renormalization \wrt to the optimal Maximum Likelihood estimator, we employ the Bundle Adjustement  (BA) framework. We use the solution of \Eq{eq:reduce_lin_solver1} in \Eq{eq:lin_solver} to compute $\lambda$'s and we then average the multiple reconstructions per point to estimate the initial points $\V{X}$. The BA algorithm minimizes the total re-projection error $\sum_i \|\V{u}_i - \hat{\V{u}}_i(.)\|^2$, where $ \hat{\V{u}}_i(.)$ is the resulting non-linear mapping of $\V{v}_0$, $\V{g}_0$ and the auxiliary variable $\V{X}$, while index $i$ runs over all the observations. Note that our goal is to use a standard framework as a baseline to evaluate the performance of the proposed estimator. Therefore, we stay with the same and necessary parameters of velocity and gravity and we do not augment the set of unknowns with the sensor biases. To refine the parameters, the Levenberg-Marquardt algorithm is used, similar to the visual BA framework of \cite{Loukaris-ICCV2005}. 

The renormalization, despite not being an optimal ML estimator, can in practical situations well replace BA, as will be demonstrated in \Sec{sec:experiments}. The accuracy of both methods is very comparable, but the computational burden differs. There are multiple advantages of the renormalization over BA, as the renormalization
\begin{itemize}
\item does not need auxiliary variables to be introduced as are the 3D points $\V{X}$ for BA. 
\item needs no initial conditions. Renormalization in its first iteration starts with the \cite{Taubin-PAMI1991} method and then iteratively renormalizes the matrices. BA needs a good starting point.
\item solves in each iteration a generalized eigenvalue problem of size $7 \times 7$ which is very fast and can be solved within microseconds. BA solves iteratively a linear system of normal equations of the matrices $\sim 450\times450$ in case $\sim 150$ feature points are tracked. Despite the sparsity of the problem, the computational time is by two magnitudes higher, and goes to milliseconds.
\item converges in no more than 2-5 iterations. BA needs typically at least 15 iterations.
\item provides the covariance matrix of $\V{v}_0$ and $\V{g}_0$ explicitly without any extra computations and this is directly encoded in the matrix $\M{M}$ in \Eq{eq:renorm_M_matrix}. BA computes the covariance matrix implicitly. 
\item provides an estimate of the noise level on the feature points in \Eq{eq:cov_noise_final}. In BA, one cannot explicitly estimate the noise level.
\end{itemize}

\subsection{Accelerometer and Gyroscope Bias}
\label{sec:biases}

Recall that we assume a calibrated device and any estimated offset has been removed from the \imu{} data. However, a small and slowly varying bias may still be present, while it can be modeled as a constant offset owing to the short integration time. For completeness, we show how the biases can be added. 

As shown in~\cite{Martinelli-IJCV2013}, a constant accelerometer bias can be modeled in a linear way. Such a bias can be likewise inserted into the solver of \Eq{eq:lin_solver}, that is, $\bm{\kappa}_{ij}$ can be replaced by  
\begin{equation}\label{eq:addAccBias}
\bm{\hat{\kappa}}_{ij} = \bm{\kappa}_{ij}  + \bm{\zeta}_{ij}\V{e}_a
\end{equation}
where
\begin{equation}\label{eq:AccBiasCoefficient}
\bm{\zeta}_{ij} = \left( \sum_{k=0}^{i-1} \beta_{k,i} \, \M{R}_k^0 - \sum_{k=0}^{j-1} \beta_{k,j} \, \M{R}_k^0 \right)\frac{{\Delta\tau}^2 }{2}~,
\end{equation}
and $\V{e}_a$ denotes the accelerometer bias. It is straightforward to show that $\bm{\zeta}_{ij}\rightarrow\mu_{ij}\M{I}$ when the system does not rotate. As a result, $\V{e}_a$ is not always identifiable, and separable from $\V{g}_0$. 

Instead, $\bm{\kappa}_{ij}$,  $\V{p}_i$ and $\V{p}_j$ depend on the gyroscope bias in a non-linear way. The small bias magnitude let us though use a first-order approximation, that is, $\bm{\kappa}_{ij}$ can be now replaced by
\begin{equation}\label{eq:addGyroBias1}
\bm{\hat{\kappa}}_{ij} \simeq \bm{\kappa}_{ij}  + \frac{\partial \bm{\kappa}_{ij}}{\partial\V{e}_\omega }\V{e}_\omega
\end{equation}
and likewise
\begin{equation}\label{eq:addGyroBias2}
\V{\hat{p}}_{i} \simeq \V{p}_{i}  + \frac{\partial \V{p}_{i} }{\partial\V{e}_\omega }\V{e}_\omega,
\end{equation}
where $\V{e}_\omega$ is the constant gyroscope bias and $\frac{\partial \bm{\kappa}_{ij}}{\partial\V{e}_\omega }$, $\frac{\partial \V{p}_{i} }{\partial\V{e}_\omega }$ are the respective Jacobians. Note that the gyroscope bias directly affects the rotation, that is, biased gyroscope data is integrated in \Eq{eq:imu_rot}. In order to compute $\frac{\partial \bm{\kappa}_{ij}}{\partial\V{e}_\omega }$, some useful properties of the exponential map, see (\cite{Forster-TRO2017}), thus leading to the following approximations
\begin{equation}\label{eq:rotatedVectorJacobianGyroBias}
\frac{\partial \left(\M{R}_i^0 \V{t}_\MM{c}^i\right)  }{\partial\V{e}_\omega } \simeq - \M{R}_i^0 [\V{t}_\MM{c}^i]_{_\times}\frac{\partial \M{R}_{i}^0}{\partial\V{e}_\omega }
\end{equation}
and
\begin{equation}\label{eq:betaJacobian}
\frac{\partial \left( \sum_{k=0}^{i-1} \beta_{k,i} \, \M{R}_k^0 \, \V{a}_k\right)}{\partial\V{e}_\omega } \simeq -\sum_{k=0}^{i-1} \beta_{k,i} \, \M{R}_k^0 \, [\V{a}_k]_{_\times}\frac{\partial \M{R}_{i}^0}{\partial\V{e}_\omega }~,
\end{equation}
where 
\begin{equation}\label{eq:rotationJacobian}
\frac{\partial \M{R}_i^0 }{\partial\V{e}_\omega } \simeq \sum_{k=0}^{i-1}{\M{R}_{k+1}^{i} \M{J}_k \Delta\tau} 
\end{equation}
with $\M{J}_k$ being the \emph{right} Jacobian of SO3 at $\bm{\omega}_{k} $ (see Eq.(8) in \cite{Forster-TRO2017}). The notation $[.]_{_\times}$ denotes the skew symmetric matrix. Based on \Eq{eq:rotatedVectorJacobianGyroBias}, the Jacobian $\frac{\partial \V{p}_{i} }{\partial\V{e}_\omega }$ can be computed by
\begin{equation}\label{eq:homogenousPointJacobianGyroBias}
\frac{\partial \V{p}_{i} }{\partial\V{e}_\omega } \simeq - \M{J}_{\C{N}}\M{R}_i^0 [\M{R}_\MM{c}^\imu{} \M{K}^{-1} \V{u}_i]_{_\times}\frac{\partial \M{R}_{i}^0}{\partial\V{e}_\omega },
\end{equation}
where $\M{J}_{\C{N}}$ is the Jacobian of the transformation $\C{N}(\V{x})$ and is given by the first block of $\M{J}^{(s3)}_{ij}$ in \Eq{eq:jacobian_s3}.

When both the biases need to be modeled, \Eq{eq:addAccBias} can be combined with \Eq{eq:addGyroBias1}, while the cross dependence of biases can be ignored.

Adding biases into the renormalization scheme by involving the above equations is rather straightforward. In short, in case of the accelerometer bias, the matrix $\M{S}$ in \Eq{eq:lin_solver} would contain three additional columns before the last column of $\bm{\kappa}$'s. The solution vector $\V{x}$ would contain the unknown $\V{e}_a$. As entries into these columns do not depend on $\V{u}_i$, nothing substantial changes. In case of the gyroscope bias, three extra columns would be again added into the matrix $\M{S}$ and the unknown $\V{e}_\omega$ into the solution vector $\V{x}$. The entries into $\M{S}$ now depend on $\V{u}_i$, see \Eq{eq:addGyroBias2}, so the matrix $\M{B}$
in \Eq{eq:reduce_lin_solver} has different form. Its partial derivative in \Eq{eq:derivative_B} needs to take into account the derivative of the matrix $\M{S}$ as well. The Jacobian in \Eq{eq:total_jacobian} changes to size $10\times 4$. If both biases are considered, the size of the Jacobian is $13\times 4$.

\cite{Kaiser-RAL2017} tested the robustness of~\cite{Martinelli-IJCV2013} against biased \imu{} readings. As far as the accelerometer bias is concerned, when it is identifiable, the initialization remains unaffected. In particular, their experiments show that even large unrealistic bias magnitudes can be well compensated. Therefore, we only expect a minor refinement through the renormalization scheme.

On the contrary, the initializer may be affected from a gyroscope bias when its magnitude is relatively large and the integration time is long (\cite{Kaiser-RAL2017}). However, the rolling-shutter camera allows short integration times and the initializer would not benefit much from modeling a gyroscope bias of low magnitude. As such, it is advised to leave estimation of the biases for the followed VIO system which considers much longer temporal window allowing to model their distributions more properly.

\section{Experiments}
\label{sec:experiments}

The proposed modeling is valid with either a monocular or a multiocular sensor. What is different though is the integration time needed to reliably initialize the state, because the reliability grows with the number of images. The stereo baseline leads to larger camera displacements, which in turn leads to better visual constraint via triangulation. For instance, given two successive stereo frames, the displacement from the current left to the next left camera is most of the times smaller than the distance between the current left and the next right camera. Considering more frames or widening their baseline means increasing integration time of \imu{} signals. This in general would be preferable, however, it means gathering more noise and making \imu{} contribution less trustworthy. A stereo setup allows for a good trade-off, to utilize visual information even when the camera displacement is small and the integration time of \imu{} signals is short. The stereo setup has significant advantage such that even in case of no motion, the stereo baseline still allows that the triangulation constraint to be effective and to correctly estimate zero velocity. We provide comparison of mono vs. stereo to support these arguments, however, we stick in our experiments to the stereo setup as being practically much more interesting and a suitable option, and a de facto gold-standard in wearable smart glasses.

\subsection{Synthetic Data}
\label{sec:syn_experiments}

\begin{figure*}[t]
  \begin{center}
   \begin{tabular}{c@{\hspace{10mm}}c}
    \scriptsize
    \psfrag{velocity distance err [m/s]}[cb][cb]{$\epsilon_{\V{v}_0}$ $[\textrm{m/s}]$}
    \psfrag{sigma noise [pxl]}[c][c]{$\sigma$ [pxl]}
     \psfrag{dst}[l][l]{\MM{ls}}
     \psfrag{dst_renorm}[l][l]{\MM{rnm}}
     \psfrag{dst_LM}[l][l]{\MM{ba}}
     \includegraphics[width=0.4\linewidth]{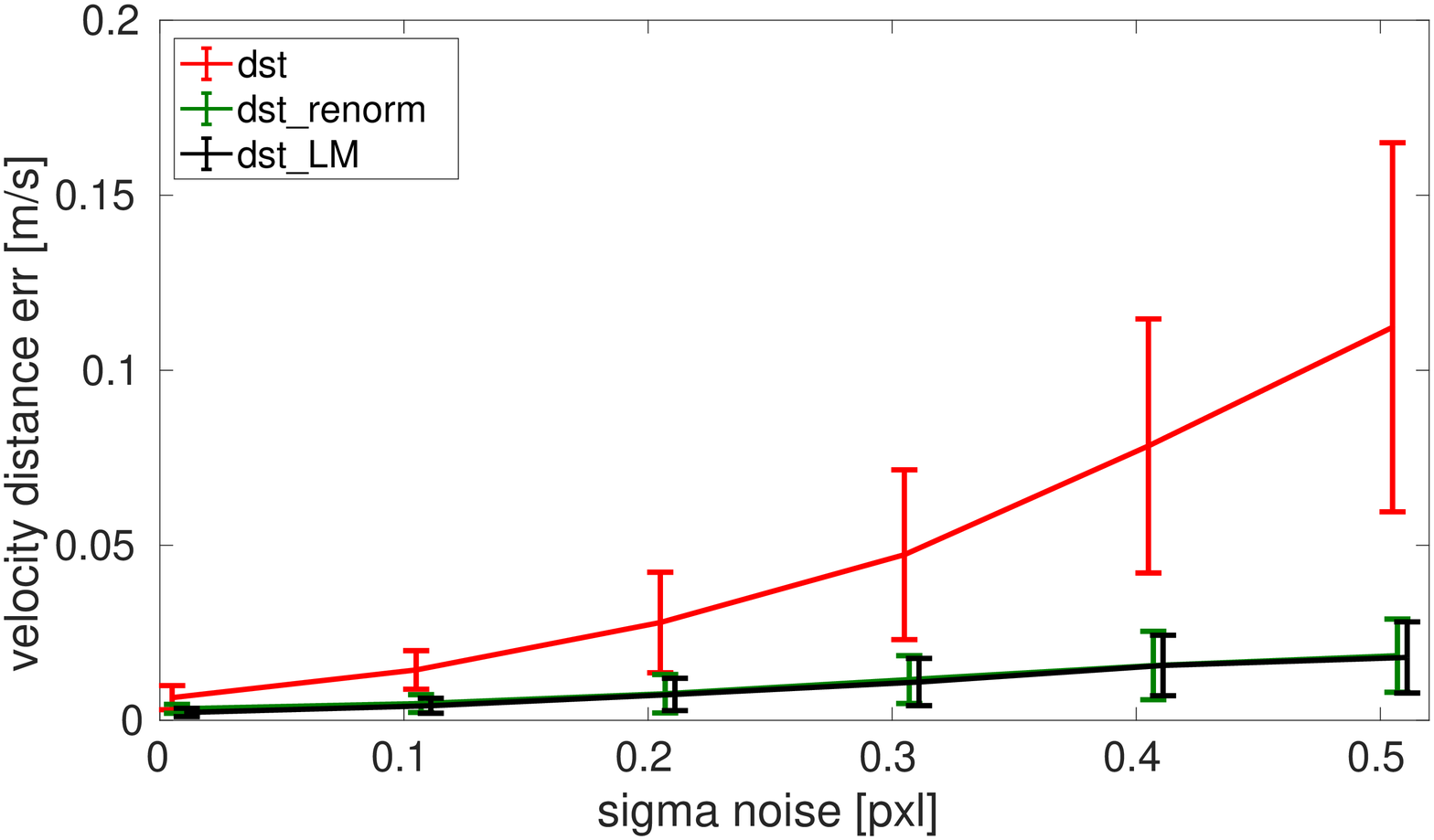} &
     \scriptsize
     \psfrag{ang gravity err [deg]}[cb][cb]{$\epsilon_{\V{g}_0}$[deg]}
     \psfrag{sigma noise [pxl]}[c][c]{$\sigma$ [pxl]}
     \psfrag{dst}[l][l]{\MM{ls}}
     \psfrag{dst_renorm}[l][l]{\MM{rnm}}
     \psfrag{dst_LM}[l][l]{\MM{ba}}
     \includegraphics[width=0.4\linewidth, trim = 1 0 0 1, clip=true ]{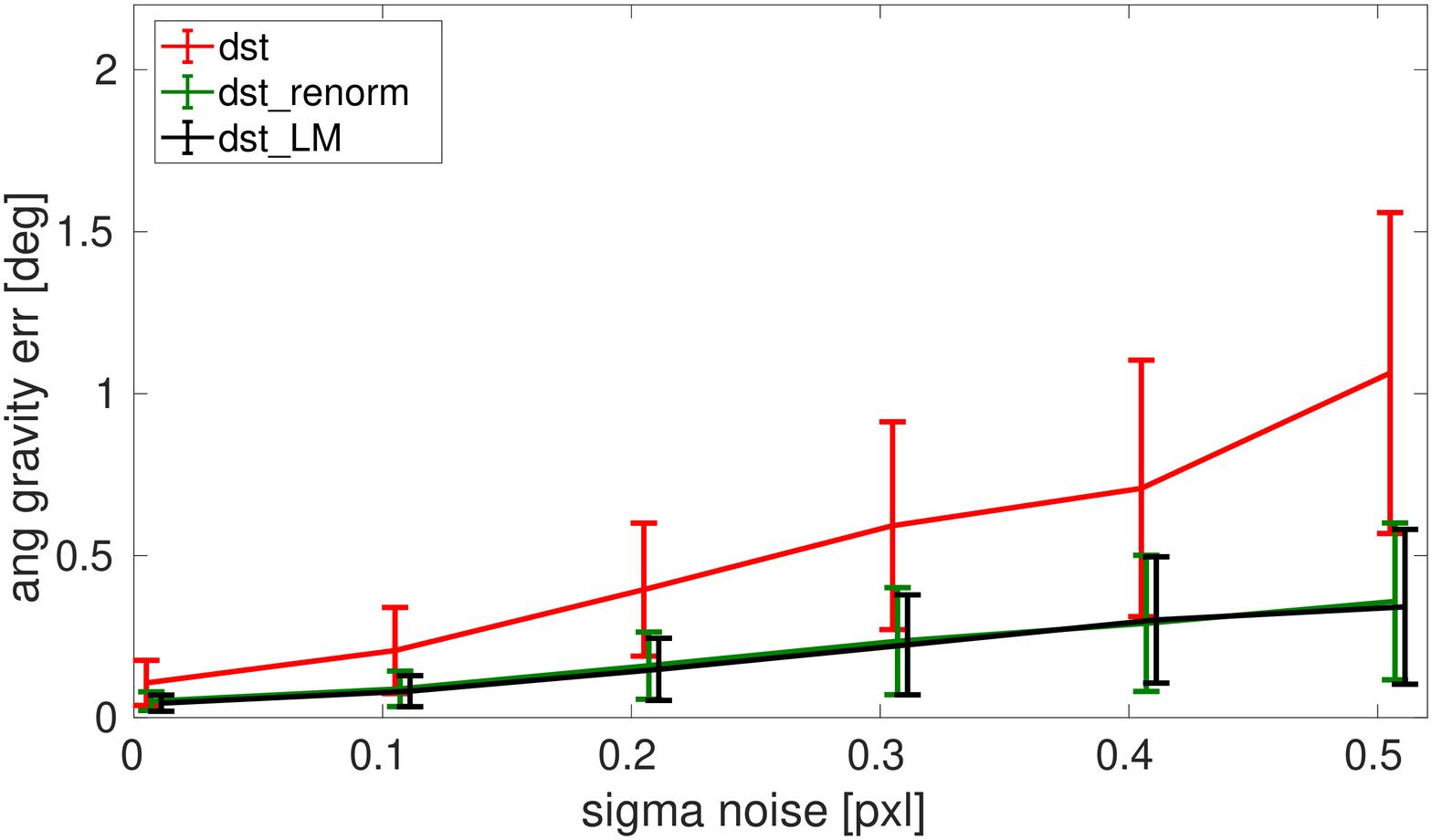} \\[-2ex]
    \end{tabular}
  \end{center}
 \caption{Synthetic noise analysis. The vertical error bars depict the standard deviation over hundreds of realizations at the particular $\sigma$ across a whole sequence. The shorter error horizontal bars on the \MM{rnm} and \MM{ba} are from the estimated covariances, while the longer ones are the empirical ones, computed as standard deviation on corresponding errors. }
 \label{fig:result_syn}
\end{figure*}

\begin{figure*}[t]
	\begin{center}
		\begin{tabular}{c@{\hspace{10mm}}c}
			\scriptsize
			\psfrag{velocity distance err [m/s]}[cb][cb]{$\epsilon_{\V{v}_0}$ $[\textrm{m/s}]$}
			\psfrag{sigma noise [pxl]}[c][c]{$\sigma$ [pxl]}
			\psfrag{LS}[l][l]{\MM{ls}}
			\psfrag{iterative reweight}[l][l]{iter reweight \MM{}}
			\psfrag{Taubin}[l][l]{Taubin}
			\psfrag{rnm}[l][l]{\MM{rnm}}
			\includegraphics[width=0.4\linewidth]{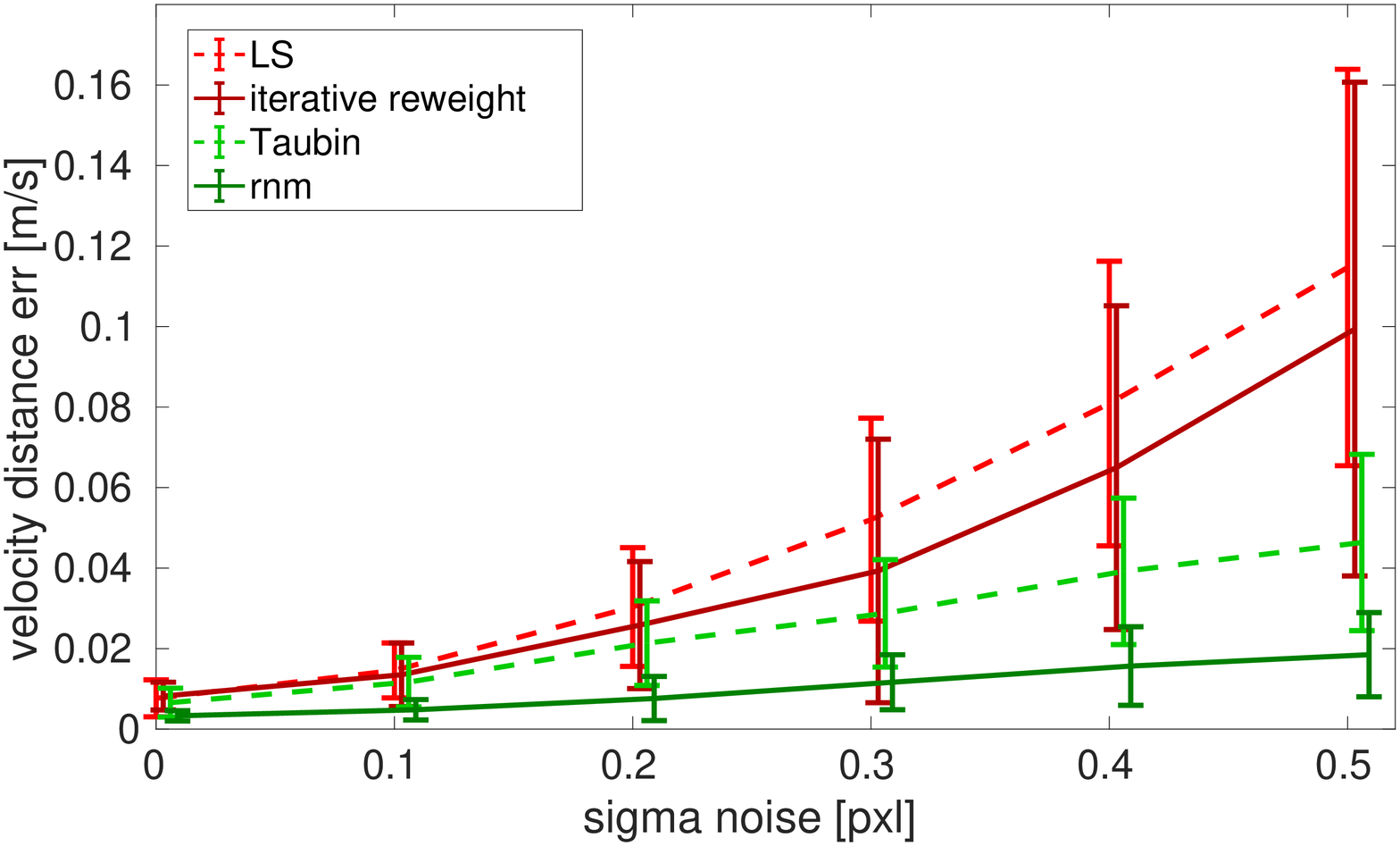} &
			\scriptsize
			\psfrag{ang gravity err [deg]}[cb][cb]{$\epsilon_{\V{g}_0}$[deg]}
			\psfrag{sigma noise [pxl]}[c][c]{$\sigma$ [pxl]}
			\psfrag{LS}[l][l]{\MM{ls}}
			\psfrag{iterative reweight}[l][l]{iter reweight \MM{}}
			\psfrag{Taubin}[l][l]{Taubin}
			\psfrag{rnm}[l][l]{\MM{rnm}}
			\includegraphics[width=0.4\linewidth, trim = 1 0 0 1, clip=true ]{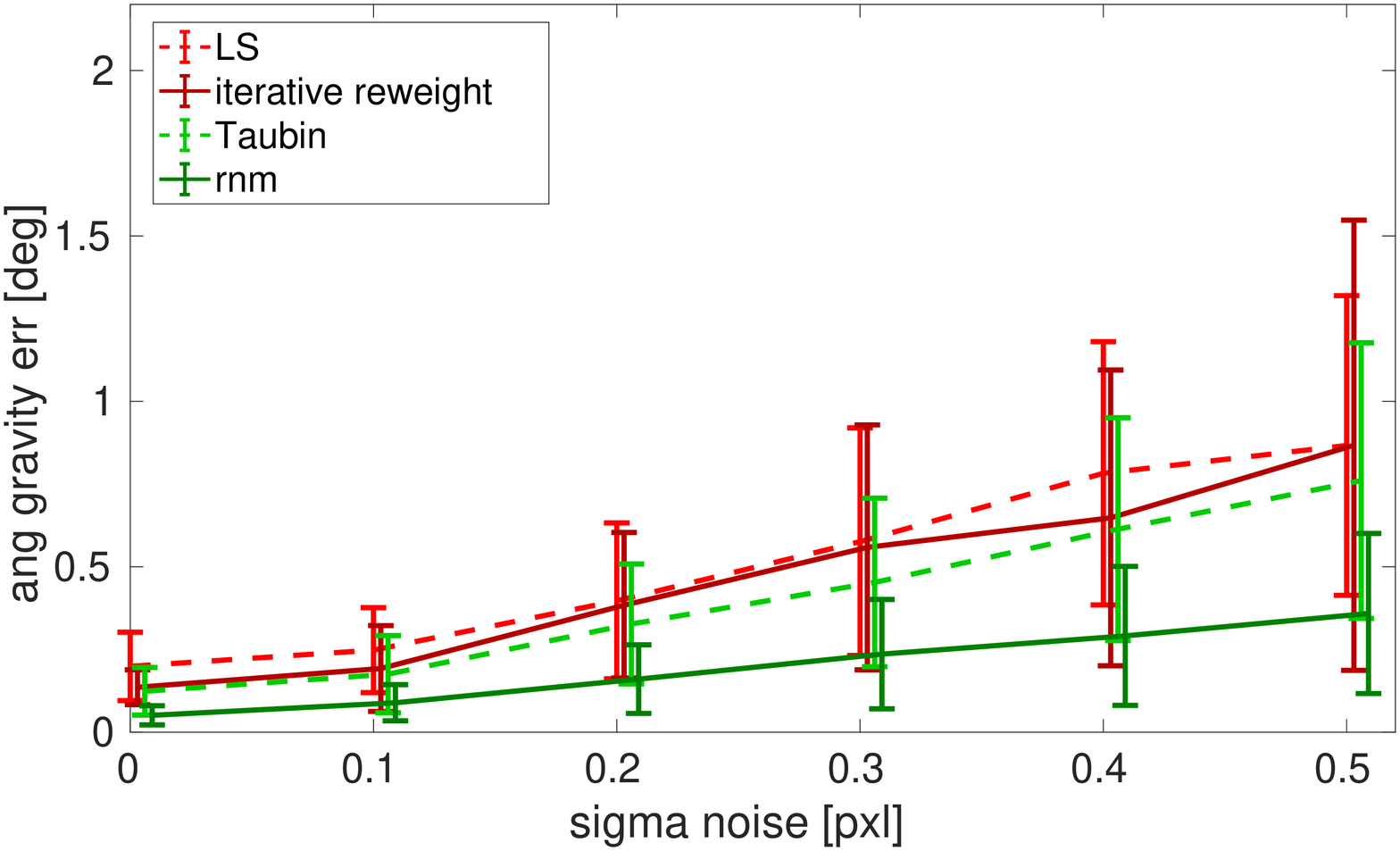} \\[-2ex]
		\end{tabular}
	\end{center}
	\caption{Comparison of the na{\"i}ve \MM{ls} method, its iterative reweight modification, Taubin method, and its iterative \MM{rnm} modification. See \Fig{fig:result_syn} for more details on axis meaning. }
	\label{fig:result_syn_Taubin}
\end{figure*}

\begin{figure*}[t]
	\begin{center}
		\begin{tabular}{c@{\hspace{10mm}}c}
			\scriptsize
			\psfrag{velocity distance err [m/s]}[cb][cb]{$\epsilon_{\V{v}_0}$ $[\textrm{m/s}]$}
			\psfrag{sigma noise [pxl]}[c][c]{$\sigma$ [pxl]}
			\psfrag{dst}[l][l]{\MM{ls (gs)}}
			\psfrag{dst_renorm}[l][l]{\MM{rnm (gs)}}
			\psfrag{dst_LM}[l][l]{\MM{ba (gs)}}
			\psfrag{dst_RS}[l][l]{\MM{ls (rs)}}
			\psfrag{dst_renorm_RS}[l][l]{\MM{rnm (rs)}}
			\psfrag{dst_LM_RS}[l][l]{\MM{ba (rs)}}
			\includegraphics[width=0.4\linewidth]{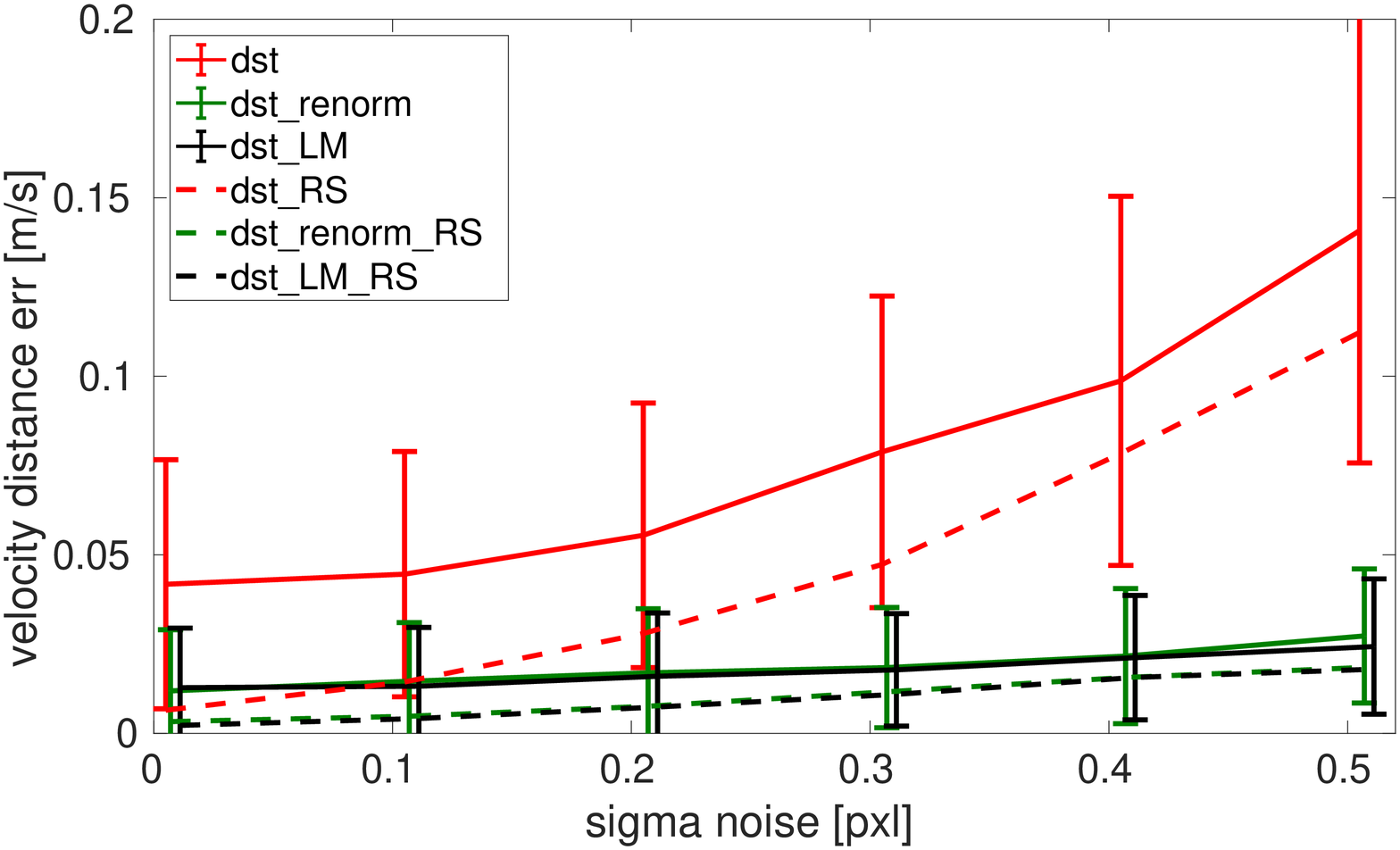} &
			\scriptsize
			\psfrag{ang gravity err [deg]}[cb][cb]{$\epsilon_{\V{g}_0}$[deg]}
			\psfrag{sigma noise [pxl]}[c][c]{$\sigma$ [pxl]}
			\psfrag{dst}[l][l]{\MM{ls (gs)}}
			\psfrag{dst_renorm}[l][l]{\MM{rnm (gs)}}
			\psfrag{dst_LM}[l][l]{\MM{ba (gs)}}
			\psfrag{dst_RS}[l][l]{\MM{ls (rs)}}
			\psfrag{dst_renorm_RS}[l][l]{\MM{rnm (rs)}}
			\psfrag{dst_LM_RS}[l][l]{\MM{ba (rs)}}
			\includegraphics[width=0.4\linewidth, trim = 1 0 0 1, clip=true ]{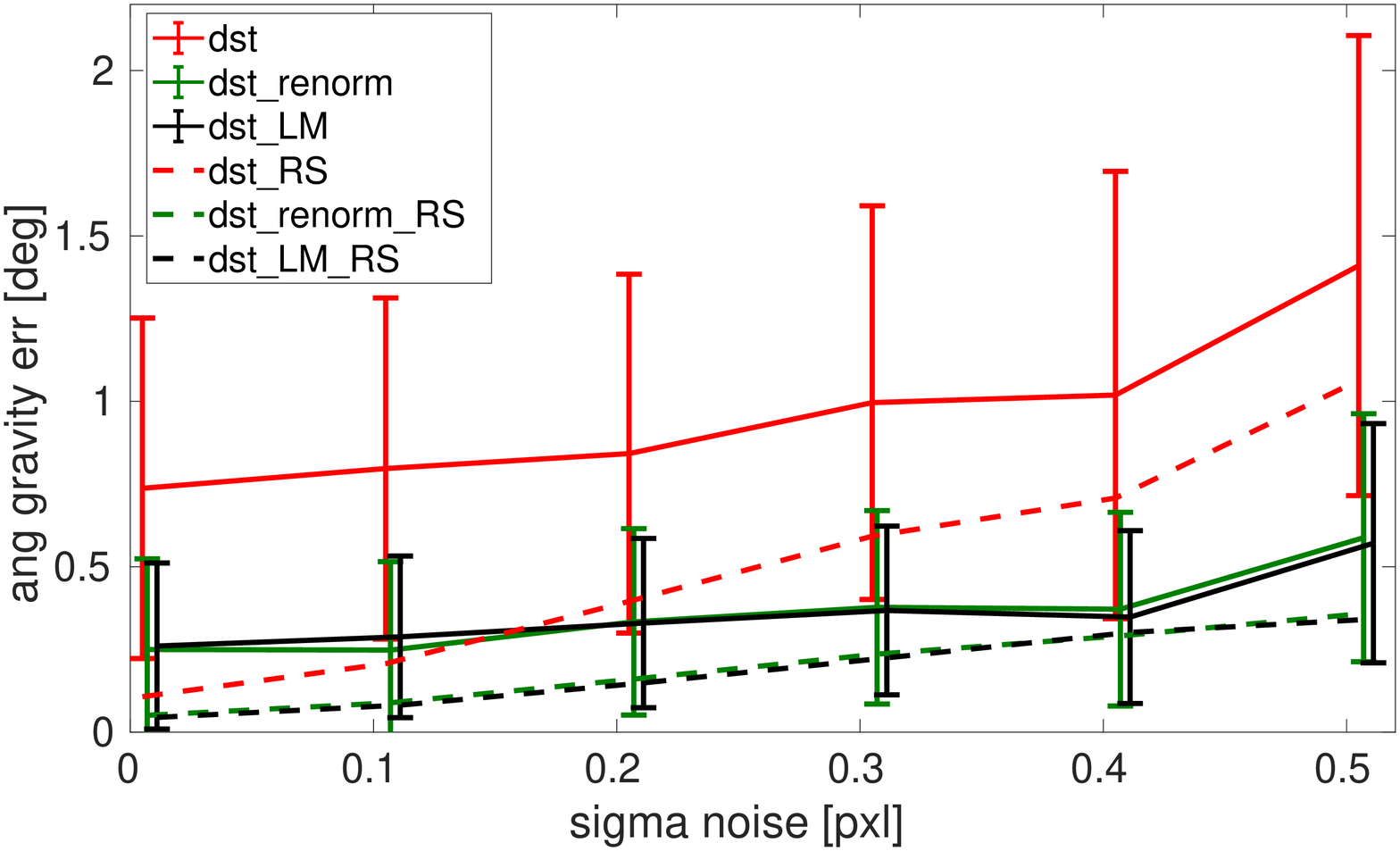} \\[-2ex]
		\end{tabular}
	\end{center}
	\caption{Effect of Global Shutter (GS) vs. Rolling Shutter (RS) camera modeling. Solid lines show the baseline \MM{ls} method of \cite{Martinelli-IJCV2013}, our proposed \MM{rnm} method, and the \MM{ba} when GS camera model is used on RS camera data. Dashed lines represent the correct modeling from \Fig{fig:result_syn}, review it for more details.}
	\label{fig:result_syn_GSvsRS}
\end{figure*}

\begin{figure*}[t]
	\begin{center}
		\begin{tabular}{c@{\hspace{10mm}}c}
			\scriptsize
			\psfrag{velocity distance err [m/s]}[cb][cb]{$\epsilon_{\V{v}_0}$ $[\textrm{m/s}]$}
			\psfrag{sigma noise [pxl]}[c][c]{$\sigma$ [pxl]}
			\psfrag{dst}[l][l]{\MM{ls (mono)}}
			\psfrag{dst_renorm}[l][l]{\MM{rnm (mono)}}
			\psfrag{dst_LM}[l][l]{\MM{ba (mono)}}
			\psfrag{dst_RSaaa}[l][l]{\MM{ls (stereo)}}
			\psfrag{dst_renorm_RSaaa}[l][l]{\MM{rnm (stereo)}}
			\psfrag{dst_LM_RSaaa}[l][l]{\MM{ba (stereo)}}
			\includegraphics[width=0.4\linewidth]{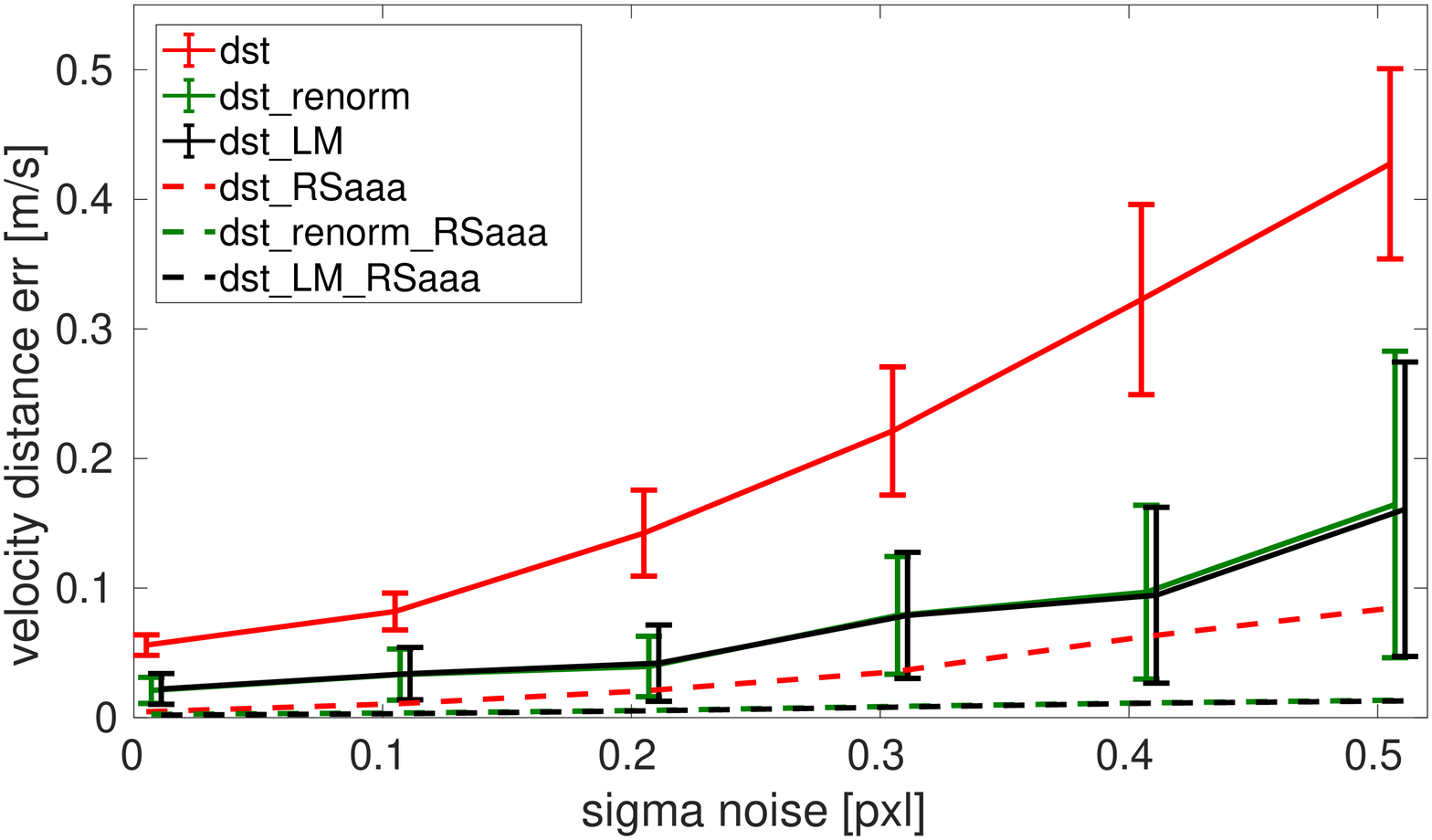} &
			\scriptsize
			\psfrag{ang gravity err [deg]}[cb][cb]{$\epsilon_{\V{g}_0}$[deg]}
			\psfrag{sigma noise [pxl]}[c][c]{$\sigma$ [pxl]}
			\psfrag{dst}[l][l]{\MM{ls (mono)}}
			\psfrag{dst_renorm}[l][l]{\MM{rnm (mono)}}
			\psfrag{dst_LM}[l][l]{\MM{ba (mono)}}
			\psfrag{dst_RSaaa}[l][l]{\MM{ls (stereo)}}
			\psfrag{dst_renorm_RSaaa}[l][l]{\MM{rnm (stereo)}}
			\psfrag{dst_LM_RSaaa}[l][l]{\MM{ba (stereo)}}
			\includegraphics[width=0.4\linewidth, trim = 1 0 0 1, clip=true ]{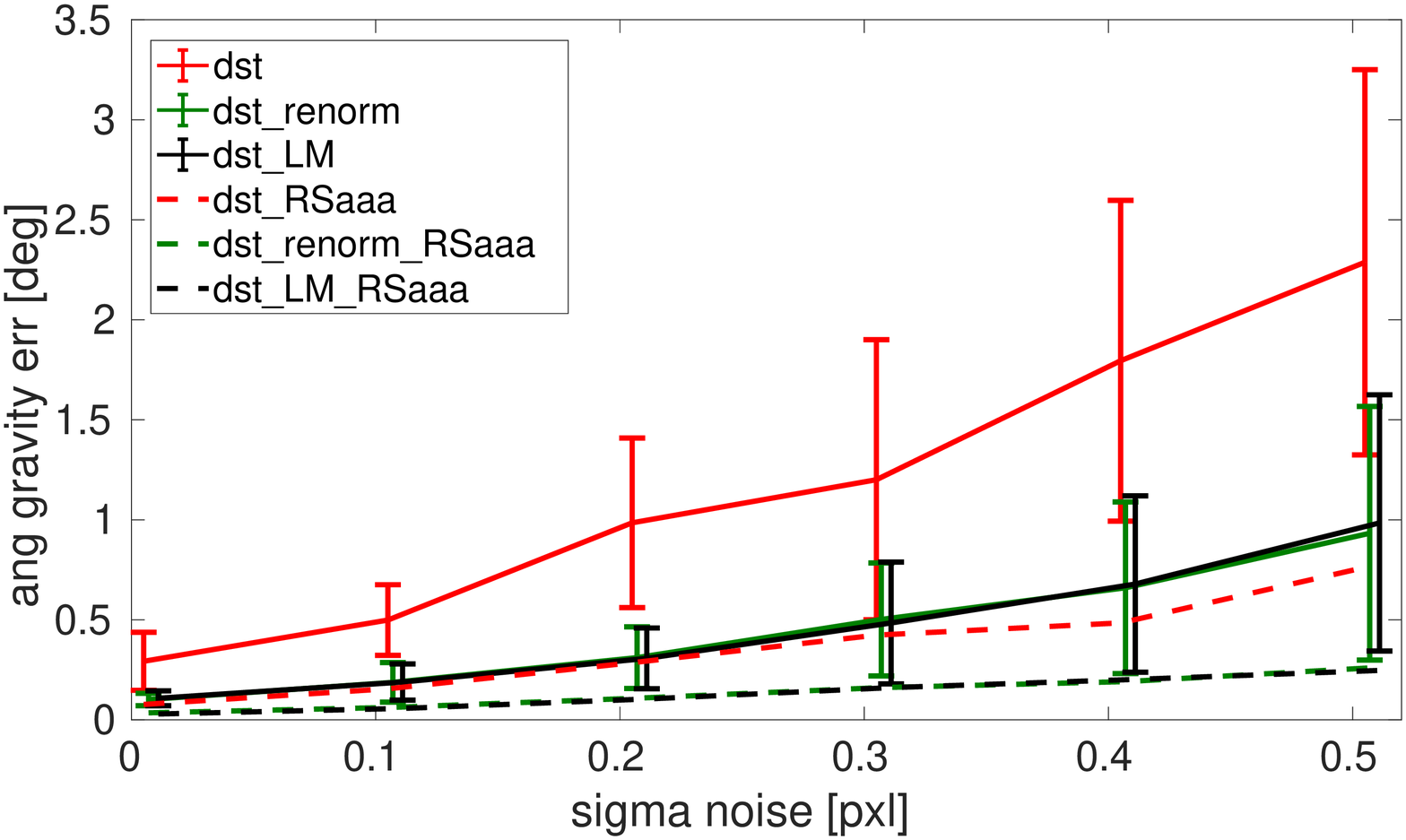} \\[-2ex]
		\end{tabular}
	\end{center}
	\caption{Mono vs. Stereo. Solid lines show the baseline \MM{ls} method of \cite{Martinelli-IJCV2013}, our proposed \MM{rnm} method, and the \MM{ba} when monocular camera is used. Dashed lines represent the stereo camera setup from \Fig{fig:result_syn}, review it for more details.}
	\label{fig:result_syn_MonoVsStereo}
\end{figure*}

In this section we perform quantitative synthetic analysis to investigate influence of noise on the final estimate of $\V{v}_0$ and $\V{g}_0$. In order to get realistic data with ground truth (GT) structure and poses, we process data from Snap Spectacles glasses with \imu{} \MM{bmi}160. States resulting from a Kalman filter on visual-inertial data play the role of GT states and high-order splines on the \imu{} data provide ideal gyroscope and acceleration readings, such that a continuous integrator perfectly interpolates between the states. The \imu{} data are then sampled at $800$Hz and finally, noise and time varying biases are added based on the calibrated variances of the used device. The device was moved forward $9$\,m along a straight trajectory with repeatedly changing viewpoint rotation from left to right. The shape of the trajectory and velocity can be seen in \Fig{fig:result_subway}(c). We simulate a stereo camera with a baseline of $14$\,cm attached to the \imu{}. During the data acquisition, the glasses were shortly static at the beginning such that we could safely initialize the \imu{} state with the static motion assumption. It allows to integrate the signals to get the ground truth poses. 
 
To produce the image correspondences,  we generate $50$ random feature points in the first left stereo image, assign them random depths in the range [1, 15]\,m and project the 3D points into the other views. We then perturb the feature points with Gaussian noise $\sigma=\{0, 0.1, \ldots, 0.5\}\,\textrm{pixels}$, the accelerometer with standard deviation of $0.005\,\textrm{ms}^{-2}$ and the rotations $\M{R}_i^0$ computed from the gyroscope data with $0.02\, \textrm{deg}$ at random orientation. At each $\sigma$ we repeat $100$ random realizations. The evaluated errors are defined as the norm on the velocity difference vector and the angle between the gravity vectors, \ie
\begin{align}
  \epsilon_{\V{v}_0}  &= \| \V{v}_0 - \V{v}_0^{\MM{gt}}\|_2, \notag\\
  \epsilon_{\V{g}_0}  &=  \angle (\V{g}_0, \V{g}_0^{\MM{gt}})
  \label{eq:error_results}
\end{align}
for one realization. We repeat the same procedure for each five-tuple of stereo images, which is slid along the whole sequence at $22$ consecutive camera positions. In all the experiments, we use consecutive five stereo cameras at $10$fps. This means five tuple of images in $0.5$s and thus the movement of $0$ - $0.3$m.To show the final statistics, at each $\sigma$ we compute mean and standard deviation. 

Recall that the visual constraints are fed into the solver in \Eq{eq:lin_solver} as image pairs, as shown for $ij$, $ik$, $kl$. Let us denote the cameras in the order: first left, first right, second left, second right and so on as $\{\MM{l}1,\MM{r}1\}$, $\{\MM{l}2,\MM{r}2\}$, $\{\MM{l}3,\MM{r}3\}$, $\{\MM{l}4,\MM{r}4\}$, $\{\MM{l}5,\MM{r}5\}$. Then we feed the following camera pairs into the matrix: $\MM{l}1-\MM{r}2$, $\MM{l}1-\MM{r}3$, $\MM{l}1-\MM{r}4$, $\MM{l}1-\MM{r}5$, $\MM{l}2-\MM{r}3$, $\MM{l}2-\MM{r}4$, $\MM{l}2-\MM{r}5$, $\MM{l}3-\MM{r}4$, $\MM{l}3-\MM{r}5$, $\MM{l}4-\MM{r}5$. We experimented with various combinations, and chose this as a trade-off between speed and accuracy. In case of a mono camera, the links would be between $\MM{l}$ cameras only.


The results can be seen in \Fig{fig:result_syn}. We compare three methods, (i) the Least Squares of \Sec{sec:min_solver}, \MM{ls}, (ii) the proposed renormalization of \Sec{sec:renorm}, \MM{rnm}, (iii) Bundle Adjustment as ML with Levenberg-Marquardt, \MM{ba}, detailed in \Sec{sec:renorm_vs_ba}, initialized by \MM{ls}. Initializing BA by \MM{rnm} rapidly speeds up the convergence, but does not improve the accuracy. For the two latter methods, \MM{rnm} and \MM{ba}, we can compute standard deviations of the estimated $\V{v}_0$ and $\V{g}_0$ from the theoretical covariance matrices. For $\MM{rnm}$, see \Eq{eq:cov_noise_final}, for \MM{ba}, see Eq.\,(A6.10) (\cite{Hartley-Book2004}). They both require knowing the noise level $\sigma$, see \Eq{eq:noise_level}. To fairly compare, we set it to the ground truth $\sigma$ at which the corresponding simulation is performed. However, we confirmed that the estimated noise level $\sigma$ by the renormalization in~\Eq{eq:cov_noise_final} is very tight to the ground truth.  As can be seen, the theoretical values are very well aligned to the empirical ones and can be well utilized in practice, to know how much to trust the final estimate. 

As expected, LS is by far the worst estimation, fairly improved by the renormalization, and very slightly polished by ML of \MM{ba}. Moreover, renormalization returns estimate of the noise level of the feature detector / tracker. Due to perturbation with the ideal Gaussian noise, the minimization of the re-projection error is a perfect Maximum Likelihood estimate. Using real data, this might be slightly violated and the \MM{ba} is not ML in its strict sense.  We will see that with the real data it may result in \MM{rnm} sometimes outperforming \MM{ba}.

{\bf LS variant methods.} We show the performance of the proposed renormalization method in comparison to previously introduced techniques which improve the na{\"i}ve LS. First, the well known iterative modification of LS (Iterative Reweight or weighted LS) is tested. The Iterative Reweight sets the matrix $\M{N}$ in \Eq{eq:gep} to identity matrix. Second, the Taubin method, where both weights $w_\alpha^{(st)}$ in \Eq{eq:renorm_M_matrix} and \Eq{eq:renorm_N_matrix} are dropped. Recall that the presented renormalization is its iterative modification. This experiment validates the claim  of \cite{Kanatani-IJCV2008}, that the error on the estimated entities can be sorted as na{\"i}ve LS $>$ weighted LS $\gg$ Taubin $>$ renormalization, see \Fig{fig:result_syn_Taubin}. Both LS and Taubin are both non-iterative methods which are interesting when computational resources are limited. The plots give intuition how much accuracy can be gained when running $5$ iterations. Each additional iteration costs the same as the iteration of the baseline method, \ie of LS for iterative reweight and of Taubin for the renormalization.

{\bf Global vs. Rolling Shutter.} We show in \Fig{fig:result_syn_GSvsRS} the systematic error, imposed by using Global Shutter camera model on Rolling Shutter camera imagery. GS camera is modeled such that a whole frame is assigned one single pose, the rotation and the translation of the middle row of the RS image. As already stated, we used trajectory reported in \Fig{fig:result_subway}(c) where the average velocity is roughly $0.7$\,m/s. Our proposed method demonstrates superior and still a reasonable performance also in this case when the camera model does not fully explain the data. In connection to VIO systems, the fact that neglecting the RS effect in camera modeling yields drift even for moderately moving walking sequences has been shown by \cite{Li-ICRA2013,Patron-Perez-IJCV2015,Schubert-ECCV2018,Schubert-IROS2019}.

{\bf Mono vs. Stereo.} We compare in \Fig{fig:result_syn_MonoVsStereo} a mono to a stereo camera case with the same frame rate, that is, the same integration time. As can be seen, the monocular case is much more sensitive to noise on the image points. Note that in common use cases a camera moves forward with epipoles being close to the image center which makes the triangulation weakly constrained. To overcome this, it would require to decrease the frame rate and thus to increase the integration time which may, however, gather too much noise. The stereo setup on the other hand keeps the visual constraint still well enforceable, independently on the motion, and provides much superior performance. Note that renormalization and BA in comparison to LS still deliver meaningful results even for the mono case, although BA needs 4 times more iterations.

{\bf More vs. Less Frames.} We show in \Fig{fig:result_syn_MonoVsStereo} comparison of more vs. less frames used for all three methods. Less frames means to still use five stereo camera frames, but considering only the pairs with the first camera only, \ie $\MM{l}1-\MM{r}2$, $\MM{l}1-\MM{r}3$, $\MM{l}1-\MM{r}4$, $\MM{l}1-\MM{r}5$. For \MM{ls} and \MM{rnm} it means $4$ frame pairs instead of $10$ which yields less entries into the input matrices $\M{S}$ and $\M{P}$ in \Eq{eq:lin_solver}. For \MM{ba} it means $5$ observations per 3D point instead of $8$. Less constraints imply lower accuracy, however, \MM{rnm} and \MM{ba} gain a speed-up of $3\times$ and $1.2\times$, respectively, for a small accuracy drop. Important note is that the proposed \MM{rnm} does not need that many observations as \MM{ls} due to the proper weighting which suppresses less confident measurements. The baseline \MM{ls} needs many more observations to statistically cancel the noise instead.

\begin{figure*}[t]
	\begin{center}
		\begin{tabular}{c@{\hspace{10mm}}c}
			\scriptsize
			\psfrag{velocity distance err [m/s]}[cb][cb]{$\epsilon_{\V{v}_0}$ $[\textrm{m/s}]$}
			\psfrag{sigma noise [pxl]}[c][c]{$\sigma$ [pxl]}
			\psfrag{dst}[l][l]{\MM{ls} (less)}
			\psfrag{dst_renorm}[l][l]{\MM{rnm} (less)}
			\psfrag{dst_LM}[l][l]{\MM{ba} (less)}
			\psfrag{dst_RSaaa}[l][l]{\MM{ls} (more)}
			\psfrag{dst_renorm_RSaaa}[l][l]{\MM{rnm} (more)}
			\psfrag{dst_LM_RSaaa}[l][l]{\MM{ba} (more)}
			\includegraphics[width=0.4\linewidth]{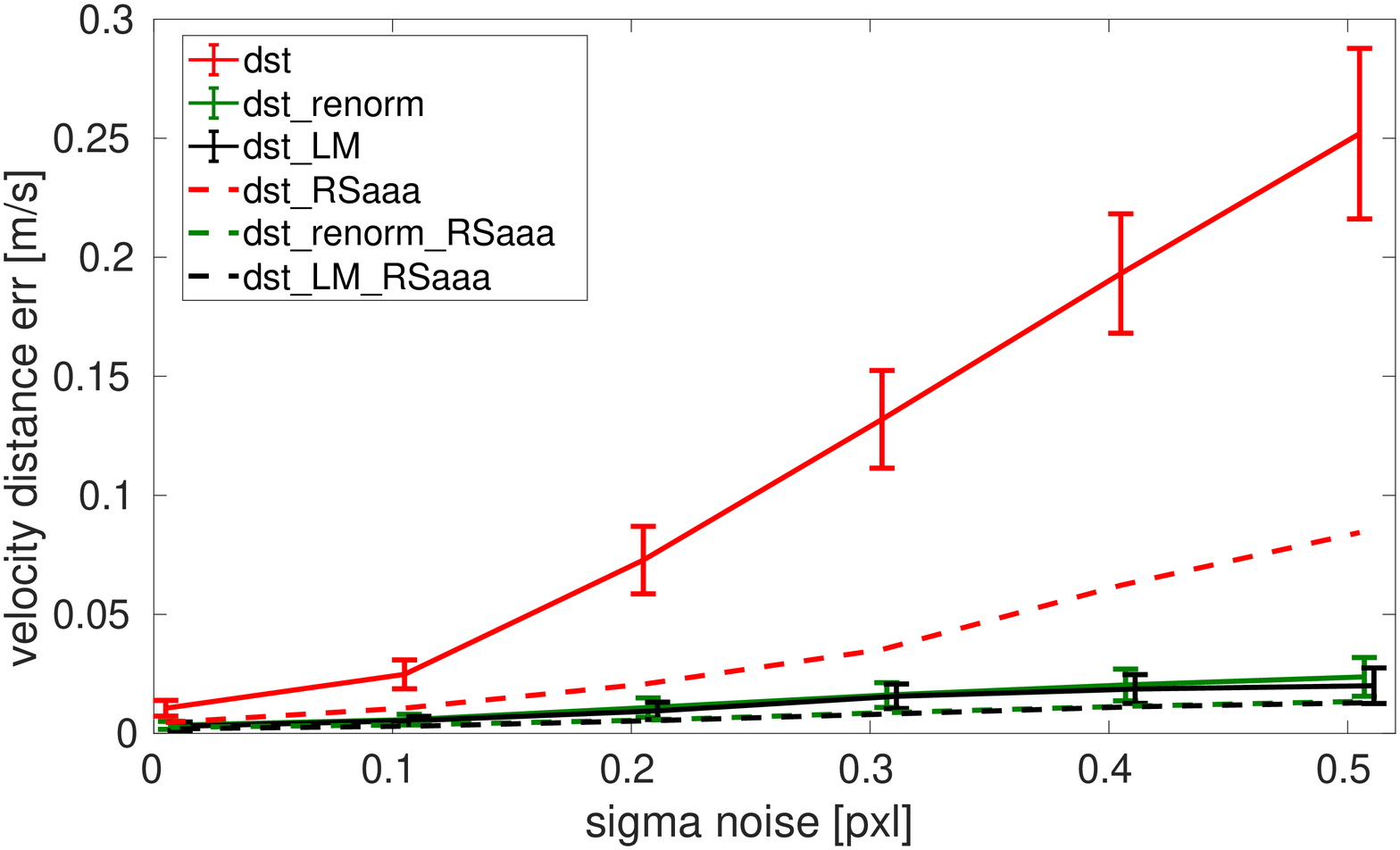} &
			\scriptsize
			\psfrag{ang gravity err [deg]}[cb][cb]{$\epsilon_{\V{g}_0}$[deg]}
			\psfrag{sigma noise [pxl]}[c][c]{$\sigma$ [pxl]}
			\psfrag{dst}[l][l]{\MM{ls} (less)}
			\psfrag{dst_renorm}[l][l]{\MM{rnm} (less)}
			\psfrag{dst_LM}[l][l]{\MM{ba} (less)}
			\psfrag{dst_RSaaa}[l][l]{\MM{ls} (more)}
			\psfrag{dst_renorm_RSaaa}[l][l]{\MM{rnm} (more)}
			\psfrag{dst_LM_RSaaa}[l][l]{\MM{ba} (more)}
			\includegraphics[width=0.4\linewidth, trim = 1 0 0 1, clip=true ]{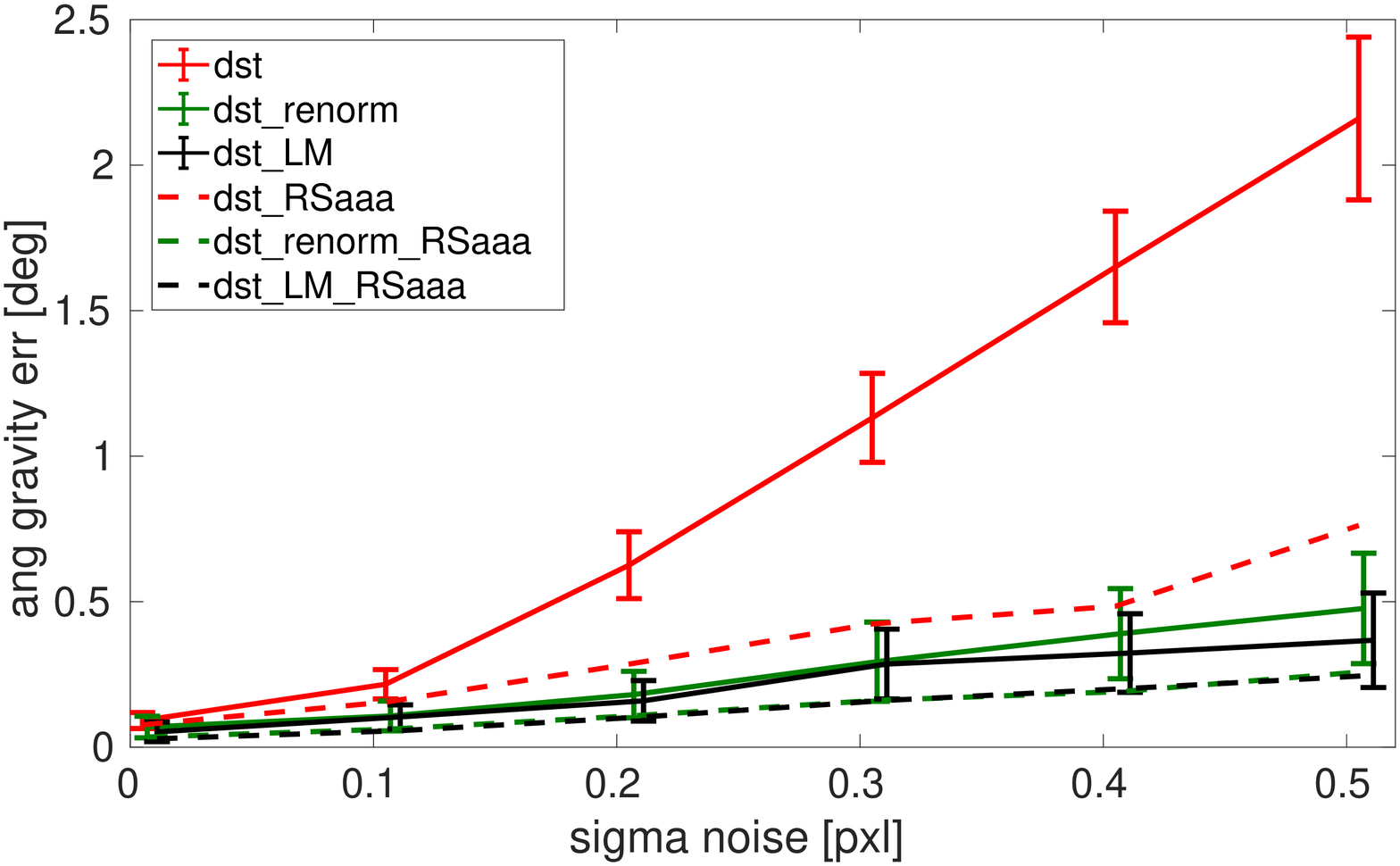} \\[-2ex]
		\end{tabular}
	\end{center}
	\caption{More vs. Less frames. Solid lines show the baseline \MM{ls} method of \cite{Martinelli-IJCV2013}, our proposed \MM{rnm} method with $4$ pairs, and the \MM{ba} when $5$ frames are used. Dashed lines represent the stereo camera setup from \Fig{fig:result_syn} with $10$ frame pairs and $8$ frames, respectively. }
	\label{fig:result_syn_MoreVsLess}
\end{figure*}

Synthetic experiments presented in this section allow us to perform noise perturbation analysis and to give the reader better intuition on different configurations. We skip such detailed comparisons for real sequences, as we see analogous behavior which leads to the same conclusions. Therefore, for following qualitative as well as quantitative results on image sequences we use the best configuration, \ie RS stereo with more frames.

\subsection{Rendered Data}
We perform qualitative comparison on realistic rendered image data, as this gives us perfect ground truth to compare to. We deploy Unreal Engine of \cite{Unreal} for rendering the images. We obtained the trajectories and the \imu{} data the same way as described in the previous section, for various types of walking trajectories of Snap Spectacles glasses. We simulate two virtual VGA rolling shutter cameras with noisy sensors, with the baseline of $14$\,cm, and readout time of $10$\,ms. As an input into the initializer, the \imu{} data is perturbed by Gaussian noise and biases on accelerometer and gyroscope with random walk noise. The features are detected by the FAST corners of \cite{Rosten-PAMI2008} and further tracked by the ECC tracker of \cite{Evangelidis-PAMI2008}. In order to prune outliers we use vanilla \MM{ransac} with the minimal solver of \Sec{sec:reduced_min_solver}. To confirm the feasibility of the statistical assumption, we plot the error distribution of the tracked features which can be obtained through known depth values of the rendered images. As the \Fig{fig:gaussian_noise} depicts, the distribution is Gaussian with subpixel accuracy. Based on this observation, we believe that Gaussian distribution on the image correspondences is a reasonable assumption.

\begin{figure}[t]
  \begin{center}
    \begin{tabular}{@{\hspace{0mm}}c@{\hspace{2mm}}c}
       \includegraphics[width=0.49\linewidth]{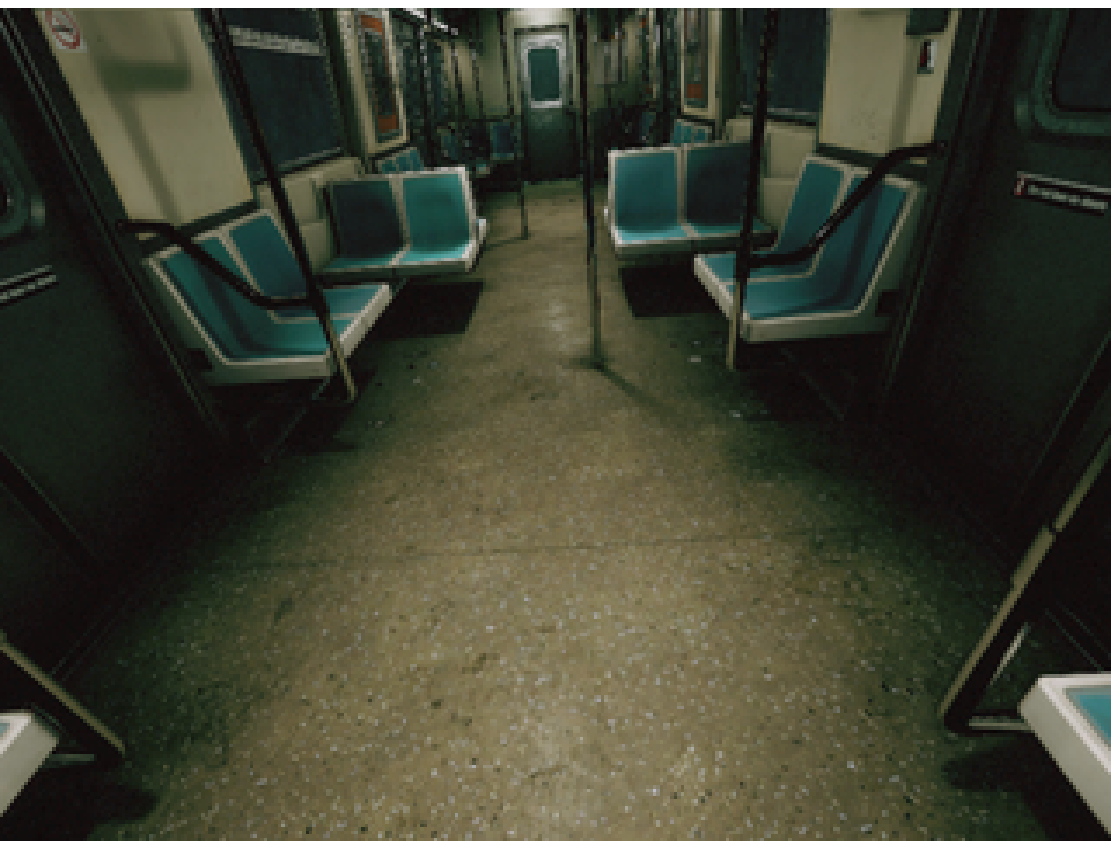} &
       \includegraphics[width=0.49\linewidth]{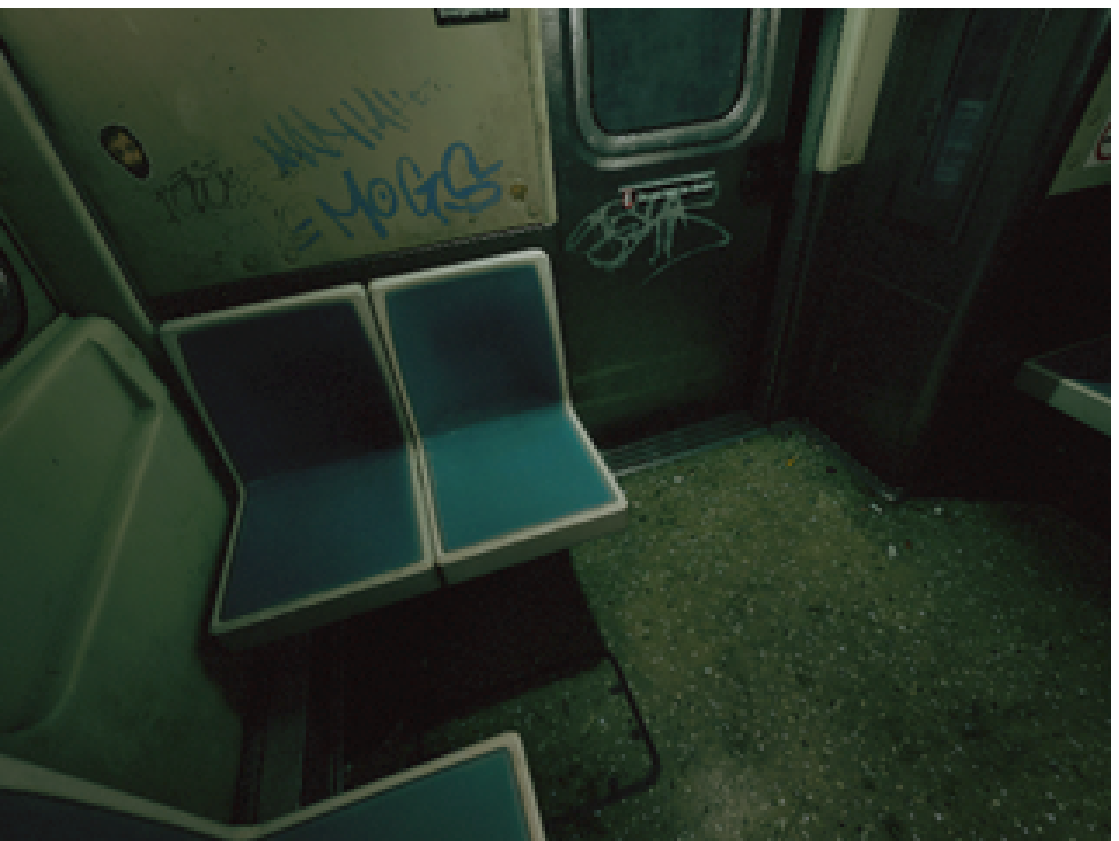} \\
       (a) & (b)\\[1ex]
     \end{tabular} 
     \scriptsize
     \psfrag{x [m]}[bc][bc]{$x [\textrm{m}]$}
     \psfrag{y [m]}[bc][bc]{$y [\textrm{m}]$}
     \includegraphics[width=0.98\linewidth, trim = 1 0 0 1, clip=true]{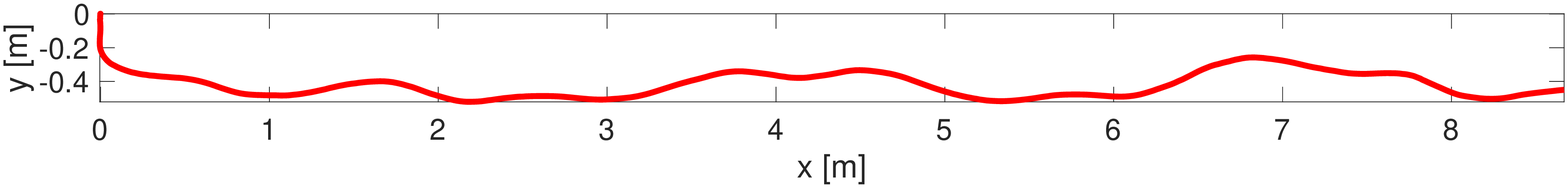} \\[2ex]
     \psfrag{velocity [m/s]}[bc][bc]{velocity [m/s]}
     \includegraphics[width=0.98\linewidth, trim = 1 0 0 1, clip=true]{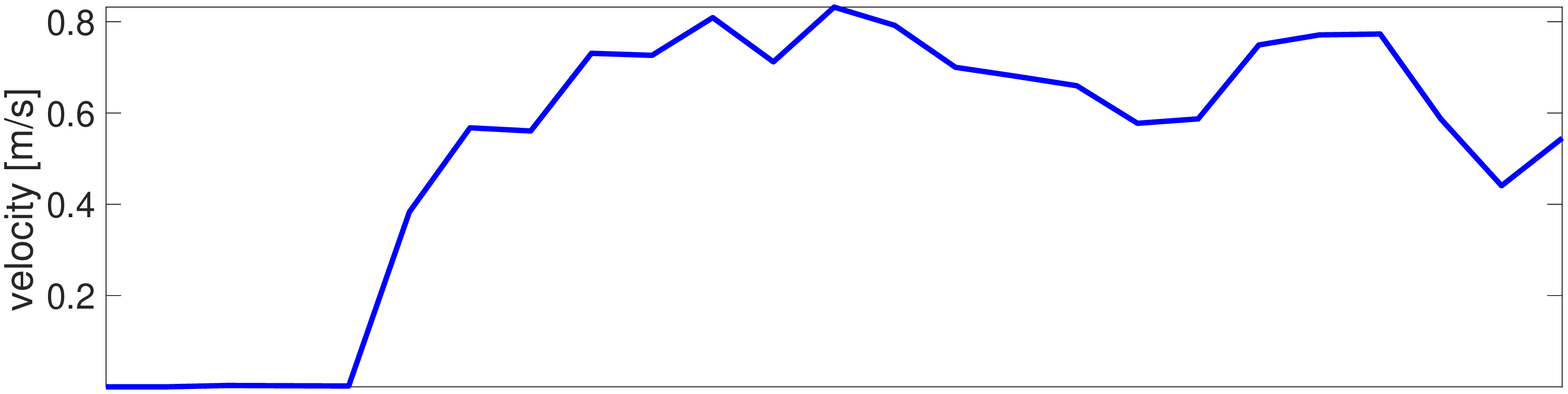}\\
     (c)\\[2ex]
    \scriptsize
    \psfrag{LS_inl}[l][l]{\MM{ls}}
    \psfrag{renorm_inl}[l][l]{\MM{rnm}}
    \psfrag{LM_inl}[l][l]{\MM{ba}}
    \psfrag{velocity dst err}[bc][bc]{$\epsilon_{\V{v}_0}$ $[\textrm{m/s}]$}
     \psfrag{batch ID}[tc][tc]{batch ID}
    \includegraphics[width=0.98\linewidth,  trim = 1 0 0 1, clip=true]{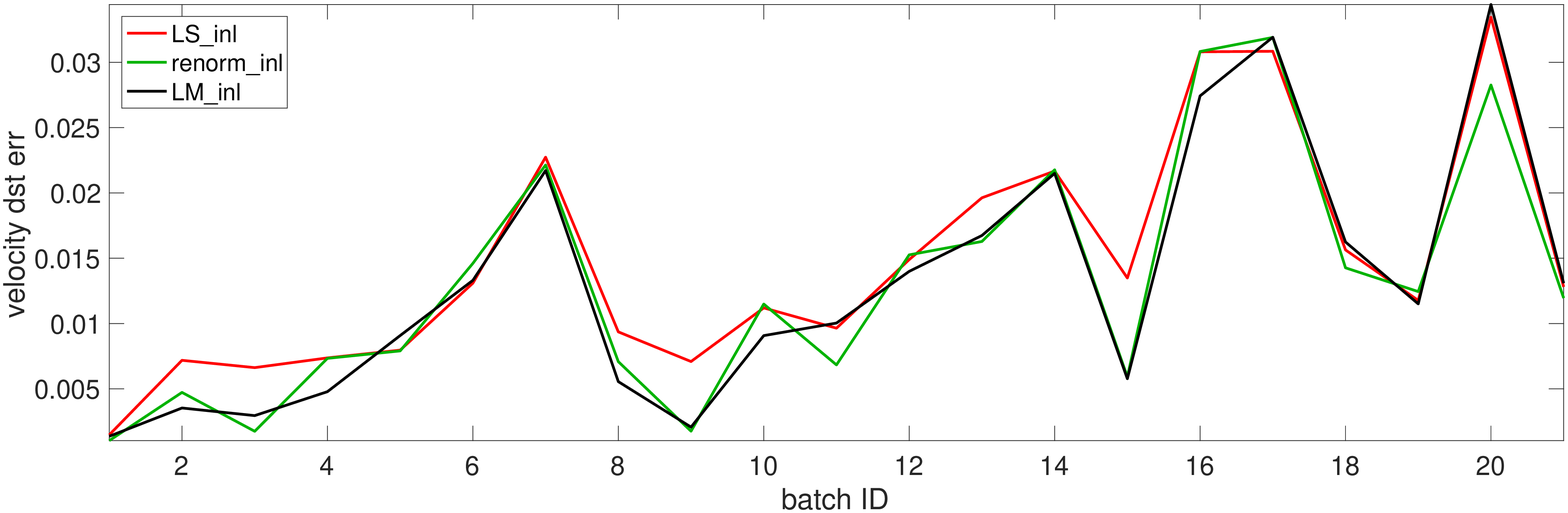} \\[2ex]
    \psfrag{gravity ang err}[bc][bc]{$\epsilon_{\V{g}_0}$[deg]}
     ~\includegraphics[width=0.98\linewidth, trim = 1 0 0 1, clip=true ]{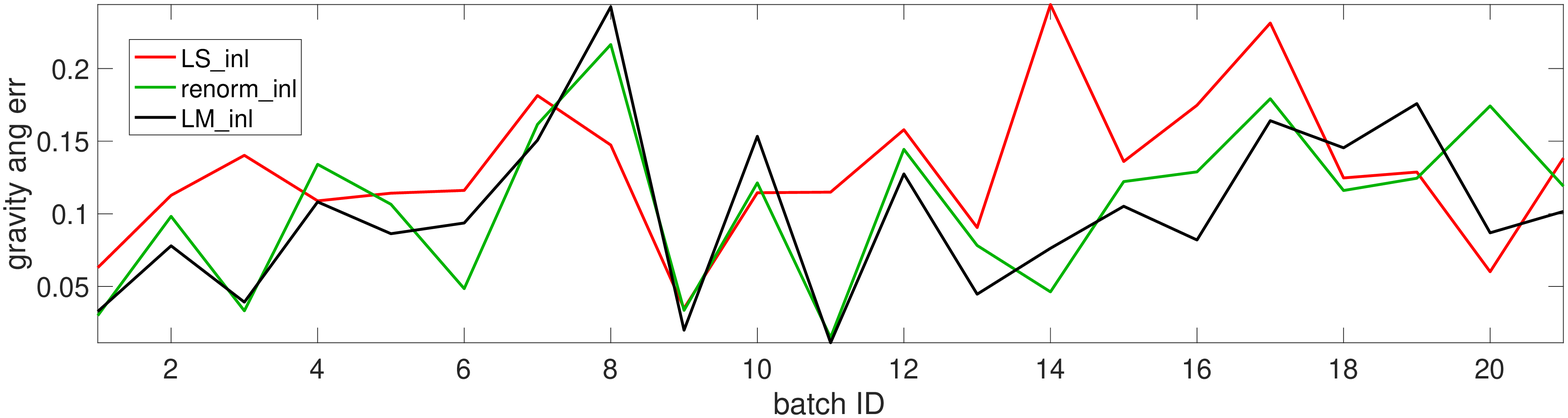} \\
     \footnotesize{(d)} \\\vspace*{-5mm}
    \end{center}
 \caption{\MM{SubwayTrain} sequence. (a)~The first and (b)~the last left stereo image of the sequence. (c)~Shape of the trajectory (top) and magnitude of the velocity during the motion (bottom). (d)~The distance error of the estimated and the ground truth initial velocity $\V{v}_0$ (top). The angular error of the estimated and the ground truth gravity $\V{g}_0$ (bottom).}
 \label{fig:result_subway}
\end{figure}

\begin{figure}[t]
  \begin{center}
    \begin{tabular}{@{\hspace{0mm}}c@{\hspace{2mm}}c}
       \includegraphics[width=0.49\linewidth]{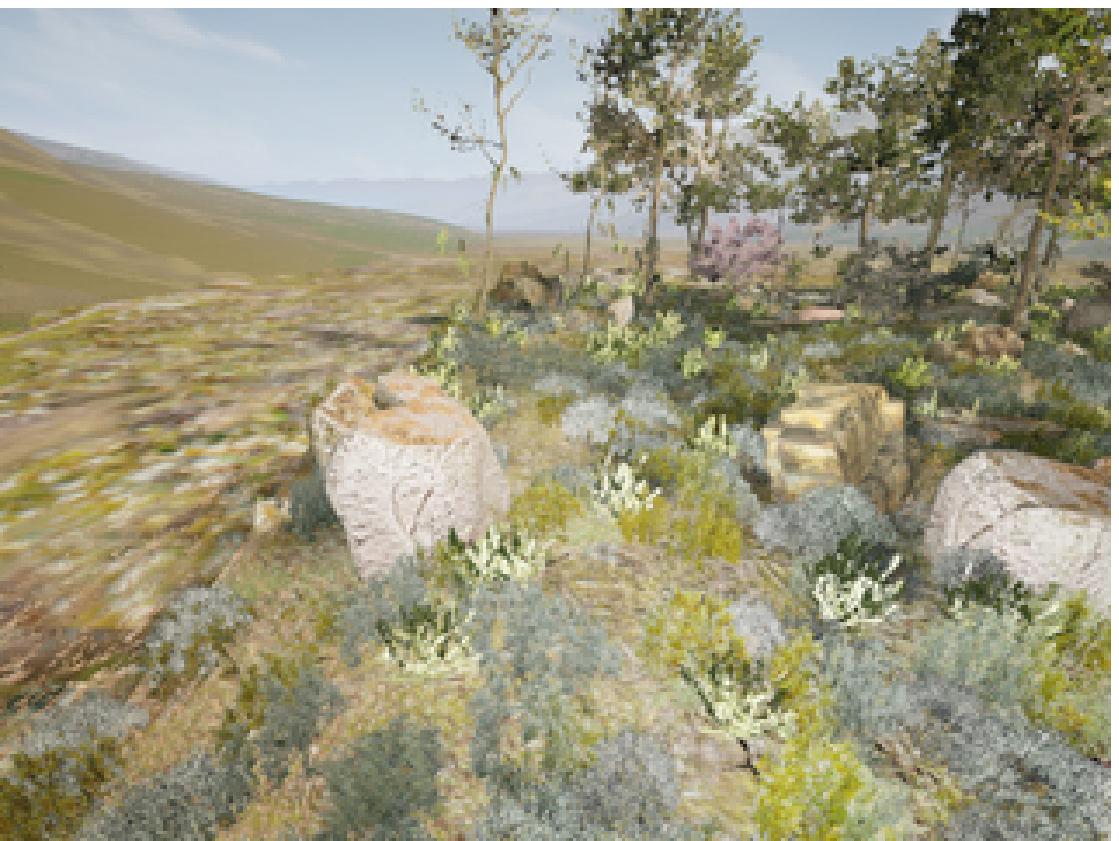} &
       \includegraphics[width=0.49\linewidth]{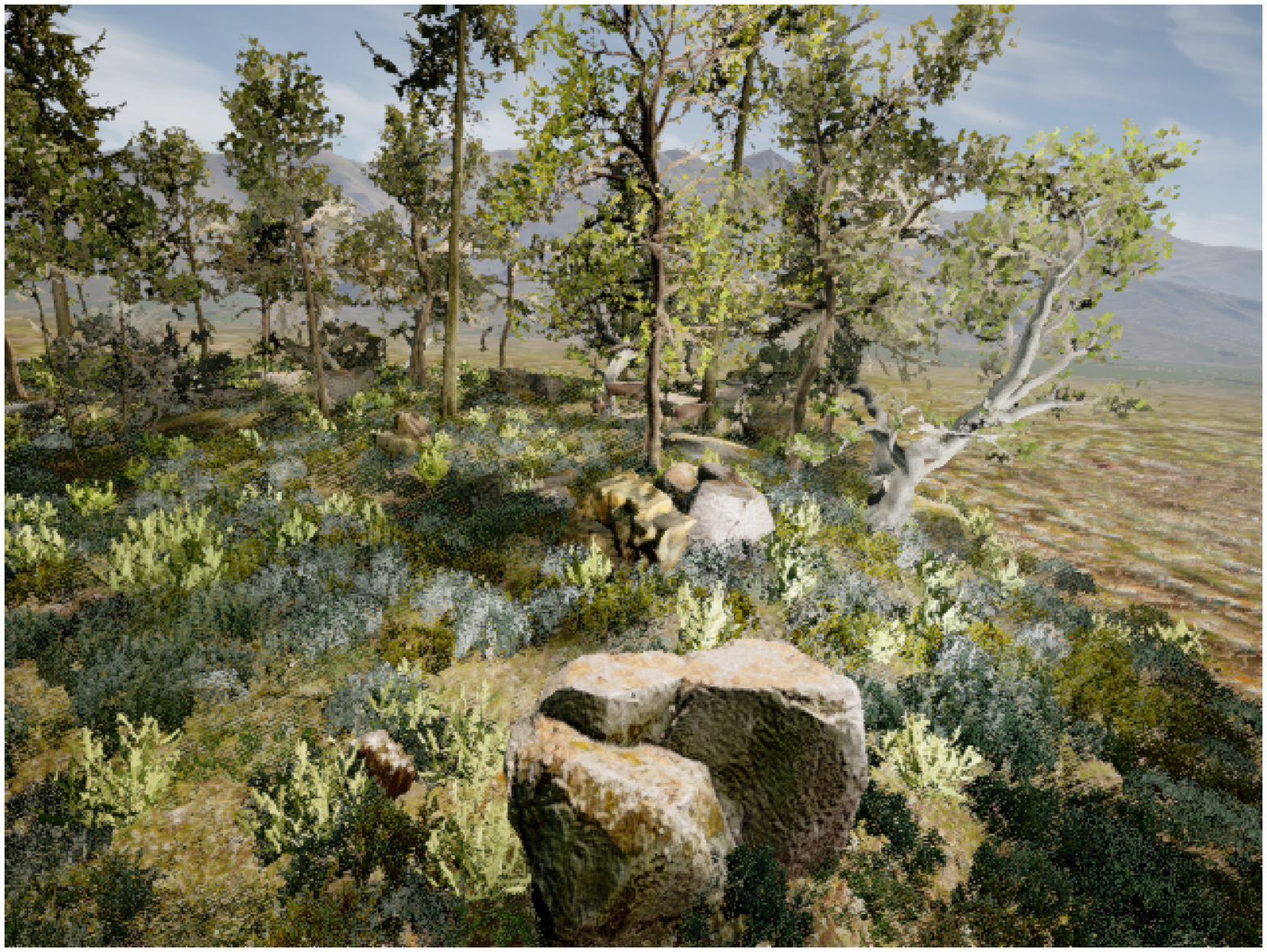} \\
       (a) & (b)\\[1ex]
     \end{tabular} 
     \scriptsize
     \psfrag{x [m]}[bc][bc]{$x [\textrm{m}]$}
     \psfrag{y [m]}[bc][bc]{$y [\textrm{m}]$}
     \includegraphics[width=0.98\linewidth, trim = 1 0 0 1, clip=true]{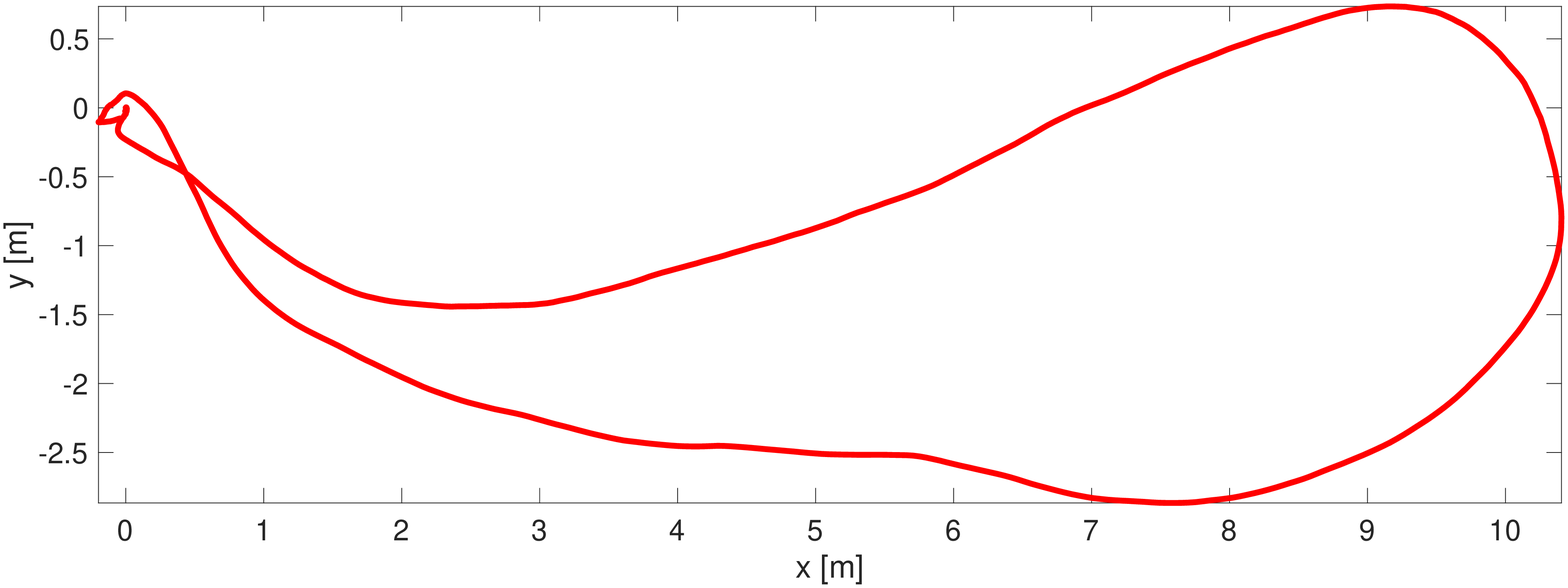} \\[2ex]
     \psfrag{velocity [m/s]}[bc][bc]{velocity [m/s]}
     \includegraphics[width=0.98\linewidth, trim = 1 0 0 1, clip=true]{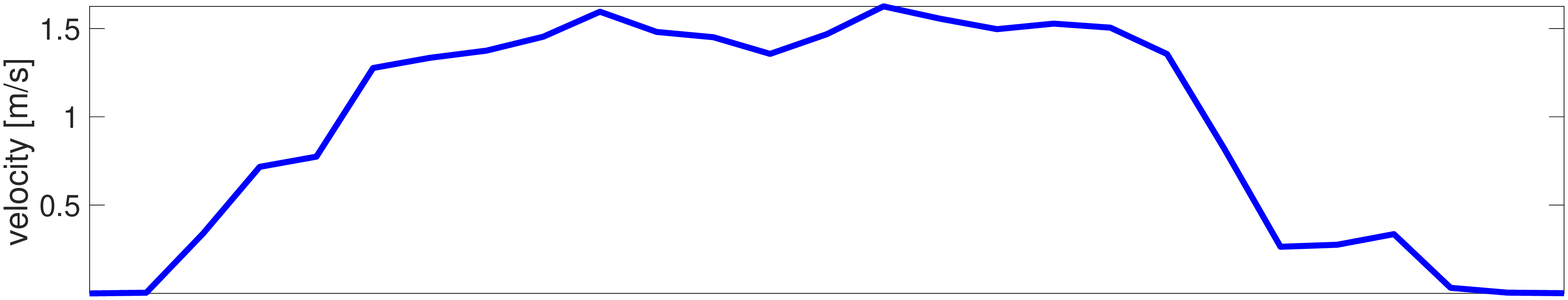}\\
     (c)\\[2ex]
    \scriptsize
    \psfrag{LS_inl}[l][l]{\MM{ls}}
    \psfrag{renorm_inl}[l][l]{\MM{rnm}}
    \psfrag{LM_inl}[l][l]{\MM{ba}}
    \psfrag{velocity dst err}[bc][bc]{$\epsilon_{\V{v}_0}$ $[\textrm{m/s}]$}
     \psfrag{batch ID}[tc][tc]{batch ID}
    \includegraphics[width=0.98\linewidth,  trim = 1 0 0 1, clip=true]{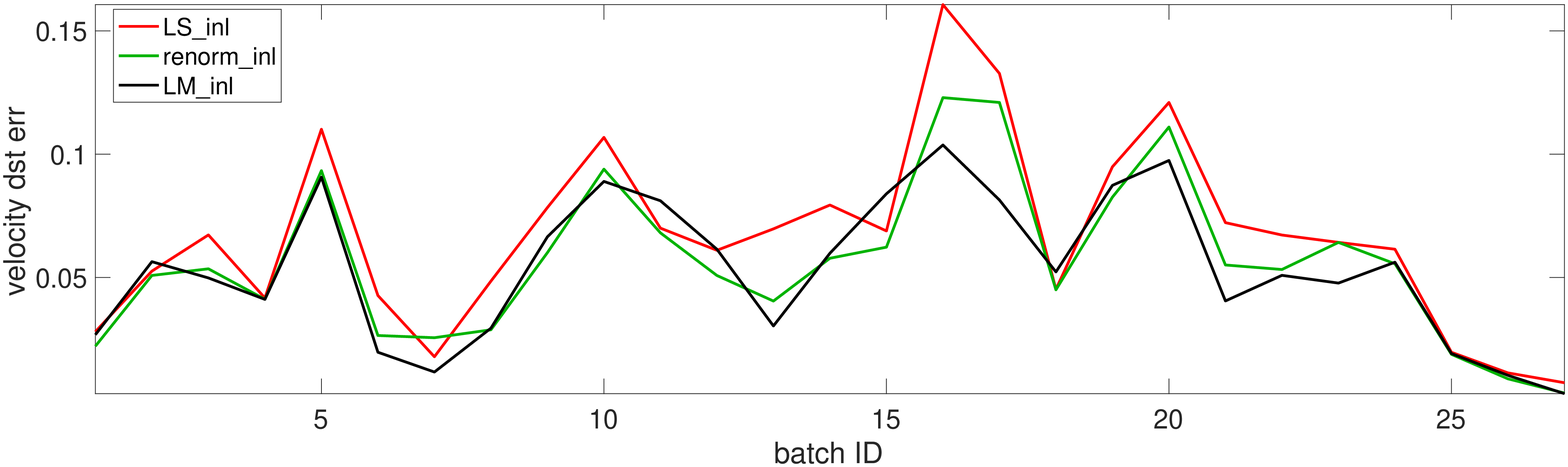} \\[2ex]
    \psfrag{gravity ang err}[bc][bc]{$\epsilon_{\V{g}_0}$[deg]}
     ~\includegraphics[width=0.98\linewidth, trim = 1 0 0 1, clip=true ]{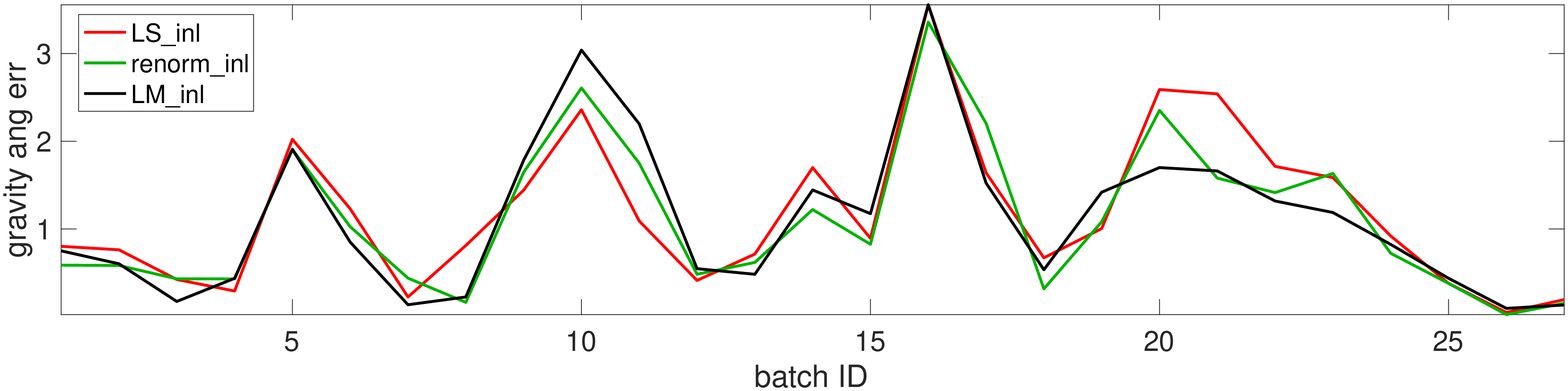} \\
     \footnotesize{(d)} \\\vspace*{-5mm}
    \end{center}
 \caption{\MM{TrapCam} sequence. Left stereo image (a)~at the start and (b)~in the middle of the sequence. See \Fig{fig:result_subway} for the remaining caption. }
 \label{fig:result_trapcam}
\end{figure}

\begin{table*}[h!]
  \begin{center}
     \begin{tabular}{|l|c|c|c|c|c|c|} 
        \hline
        & \MM{SubwayTrain}  &  \MM{TrapCam} & \MM{StorageHouse} &  \MM{StorageHouse} & \MM{SeaSideTown} & \MM{SpaceStation}\\
        & {\scriptsize forward} & {\scriptsize loop} & {\scriptsize fast shaking} & {\scriptsize loop} & {\scriptsize forward} & {\scriptsize for/back-ward}\\
      \hline\hline
       \MM{ls}  {\scriptsize [$\textrm{ms}^{-1}$ / deg]}    &     .015 / .125             &   .059 / .81.       &        .089 / 1.28          &     .027 / .47          &  .057 / 1.01 & .065 / .38\\
       \MM{rnm} {\scriptsize [\% / \%] }&    {\bf 15} /  ~5               & 	~9 / {\bf ~8}        &       {\bf 24} / {\bf ~6}   &       35 / 12 	  &  {\bf 16} / 15   & {\bf 26} /  ~7  \\
       \MM{ba} {\scriptsize [\% / \%]}  &      {\bf 15} / {\bf 33}     & 	{\bf 18} / ~5	  & 	16 / ~5 	        &      {\bf 42} / {\bf 21}   & 	~8 / {\bf 17}     &     22  /  {\bf 12} \\
        \hline
    \end{tabular}
  \end{center}
   \caption{Quantitative results. Each column shows the name of the sequence, type of the motion, mean of the absolute distance error on the initial velocity / mean of the angular error on the gravity to the baseline Least Squares method \MM{ls} in \Eq{eq:error_results}. For the \MM{rnm} and \MM{ba} improvements in percentage are shown.}
    \label{tab:results}
\end{table*}

\begin{figure*}
  \begin{center}
    \begin{tabular}{ccc}
     \includegraphics[width=0.22\linewidth]{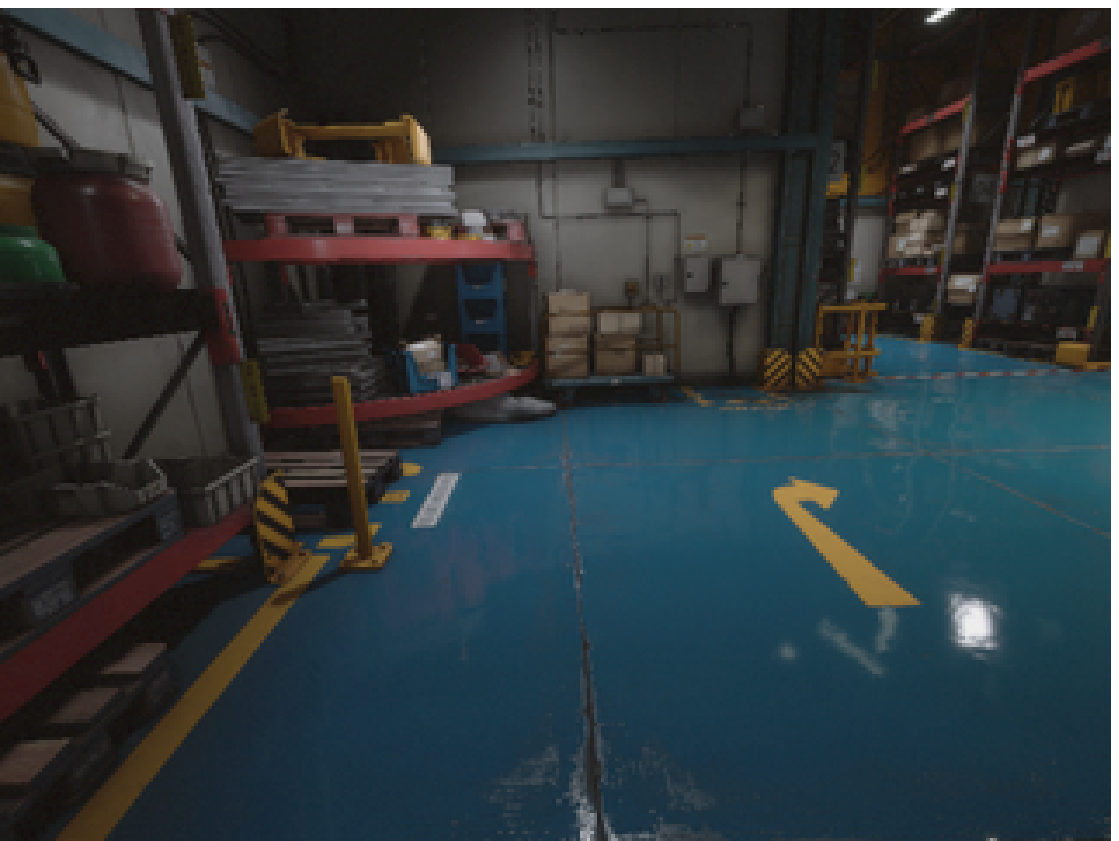} &
     \includegraphics[width=0.22\linewidth]{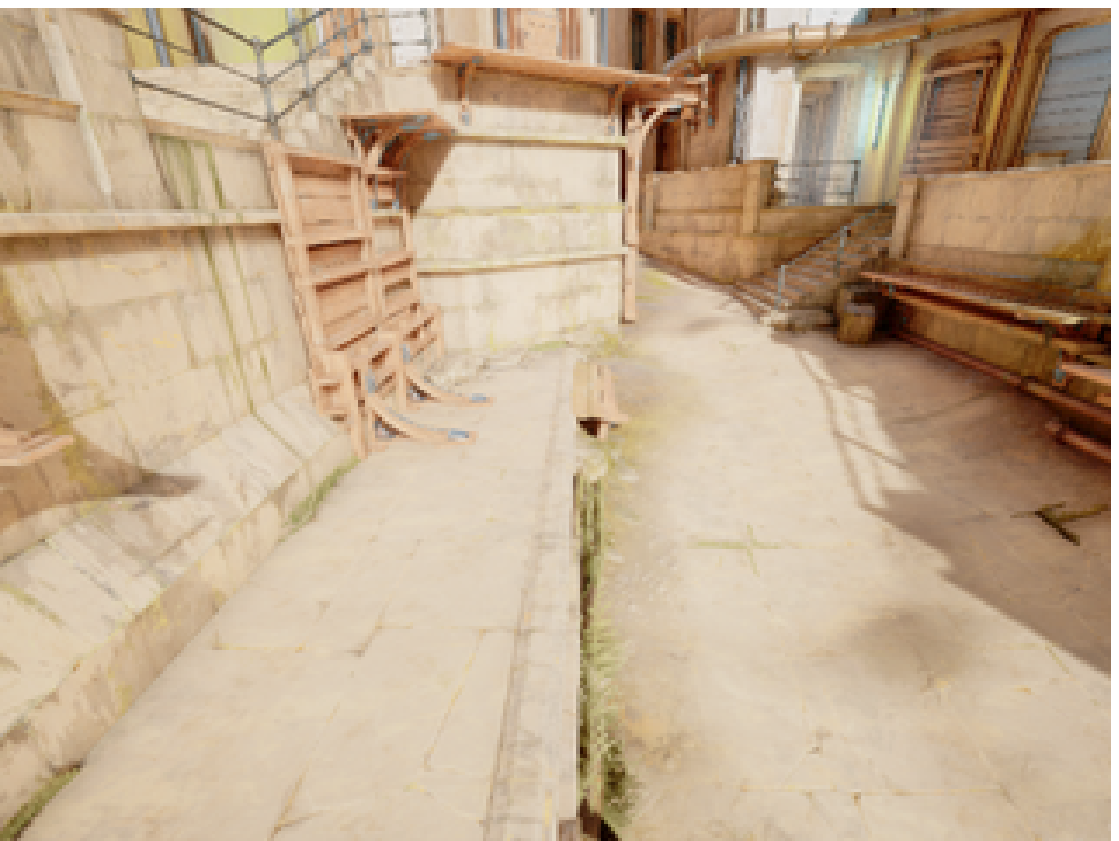} &
     \includegraphics[width=0.22\linewidth]{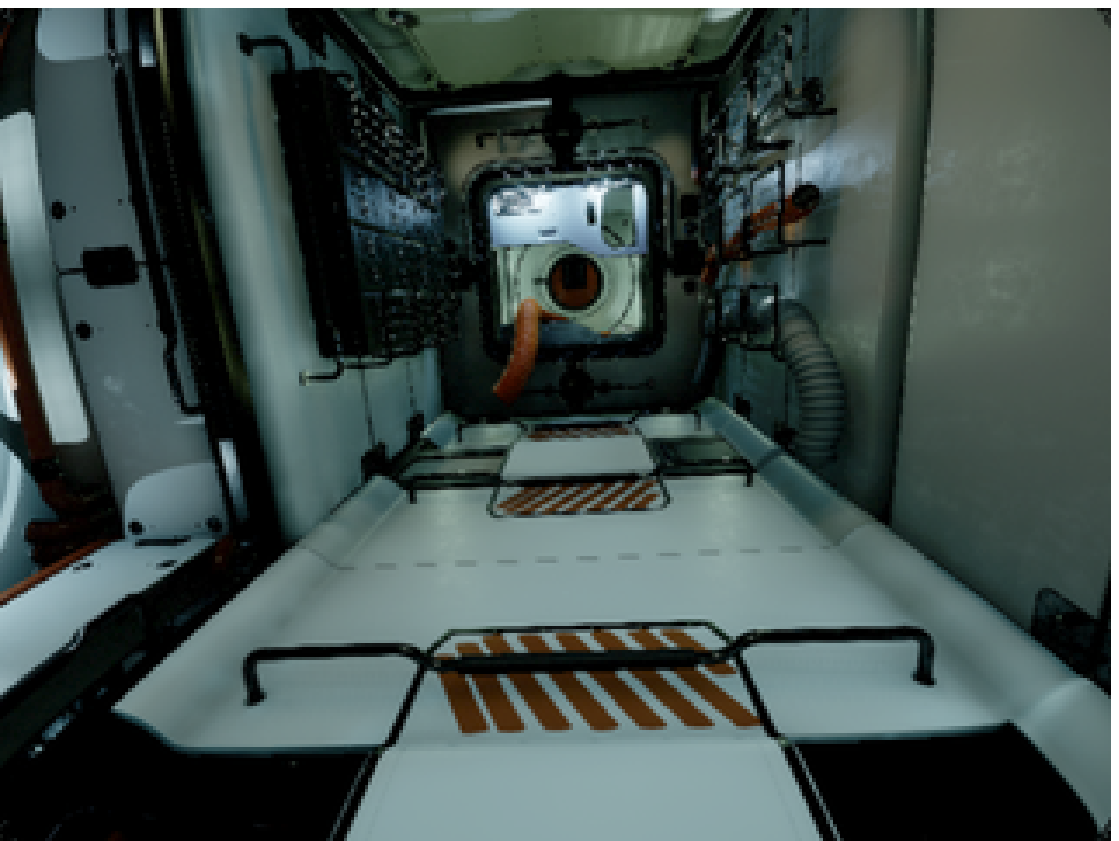} \\
     \MM{StorageHouse} & \MM{SeaSideTown} & \MM{SpaceStation}
    \end{tabular}   
   \end{center} 
  \caption{Example images of the rendered sequences.}
  \label{fig:snapshots_rendered_seqs}
\end{figure*}

\begin{figure}[t]
  \begin{center}
      \begin{tabular}{@{\hspace{0mm}}c@{\hspace{2mm}}c}
         \includegraphics[width=0.49\linewidth]{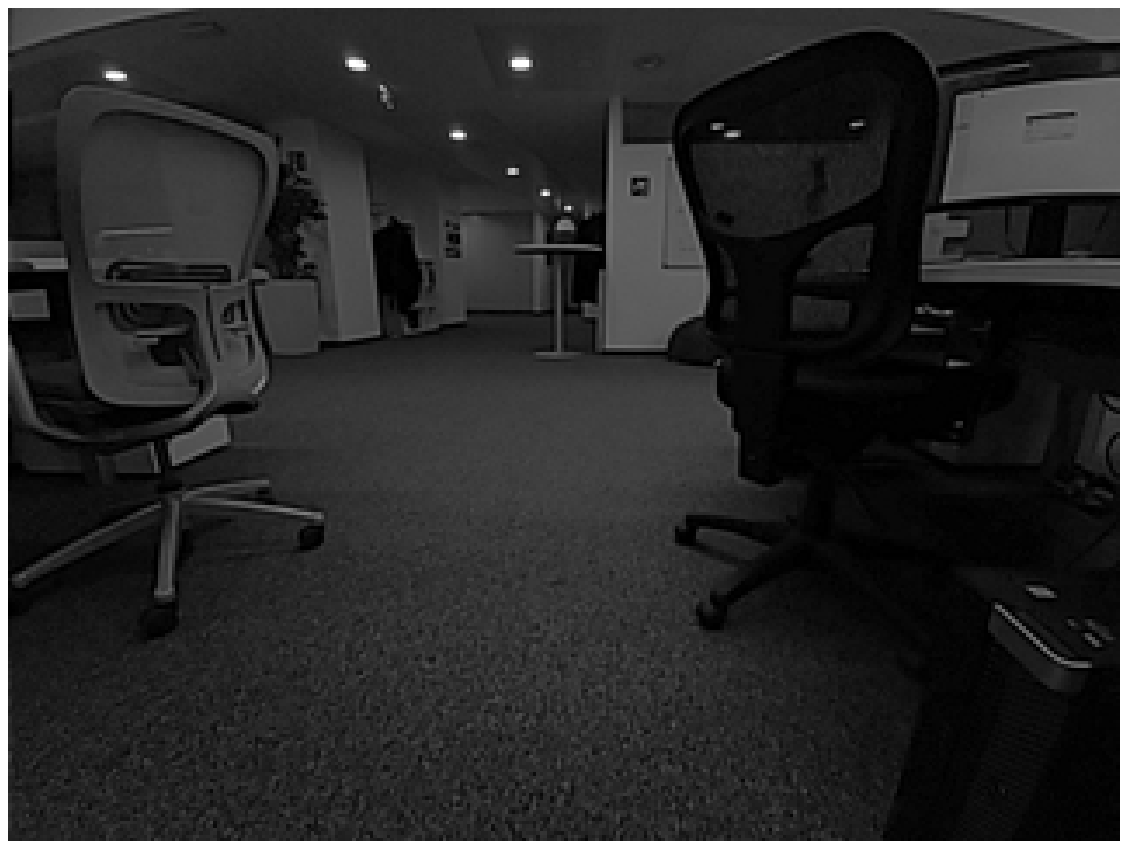} &
         \includegraphics[width=0.49\linewidth]{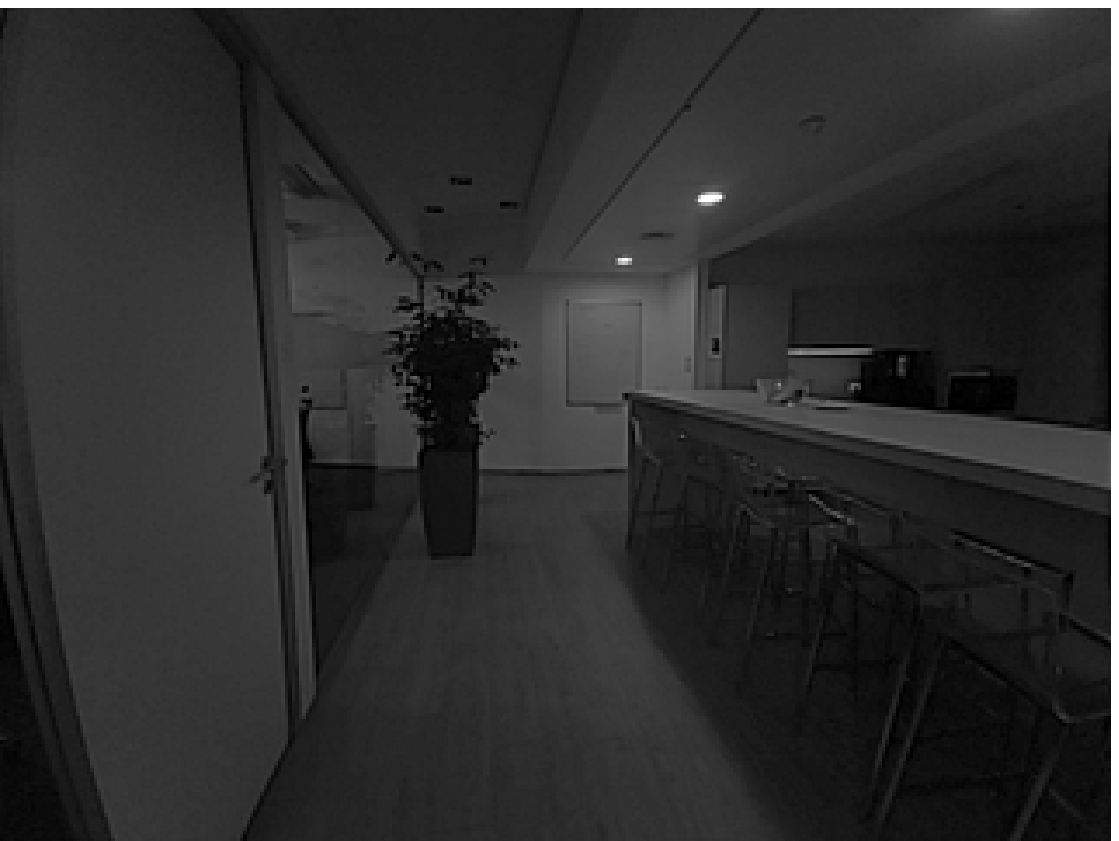} \\
         (a) & (b)\\[1ex]
     \end{tabular} 
    \scriptsize
    \psfrag{LS_inl}[l][l]{\MM{ls}}
    \psfrag{renorm_inl}[l][l]{\MM{rnm}}
    \psfrag{LM_inl}[l][l]{\MM{ba}}
    \psfrag{velocity dst err}[bc][bc]{$\epsilon_{\V{v}_0}$ $[\textrm{m/s}]$}
     \psfrag{batch ID}[tc][tc]{batch ID}
    \includegraphics[width=0.98\linewidth, trim = 1 0 0 1, clip=true]{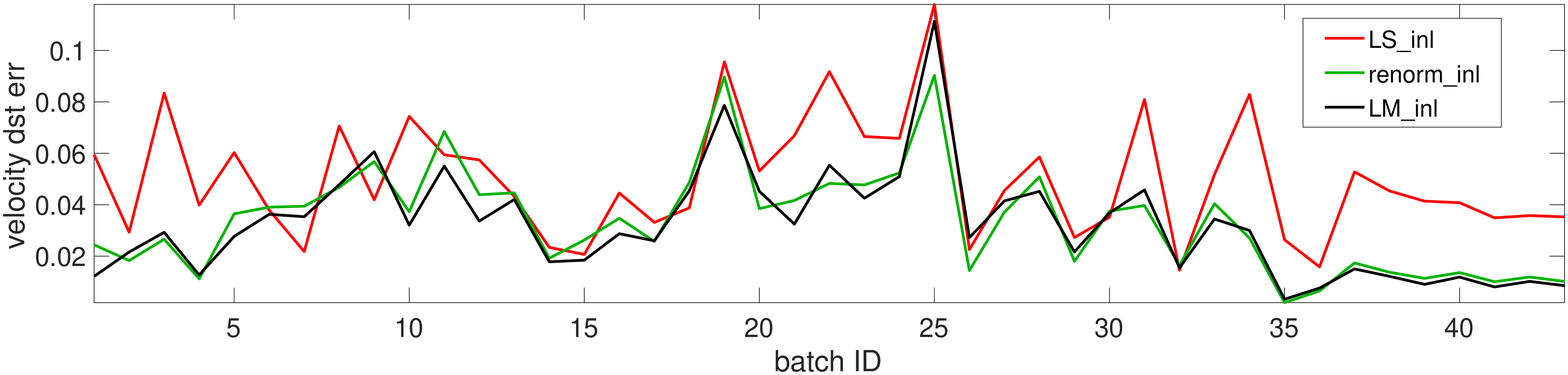} \\[2ex]
    \psfrag{gravity ang err}[bc][bc]{$\epsilon_{\V{g}_0}$[deg]}
     ~\includegraphics[width=0.98\linewidth, trim = 1 0 0 1, clip=true]{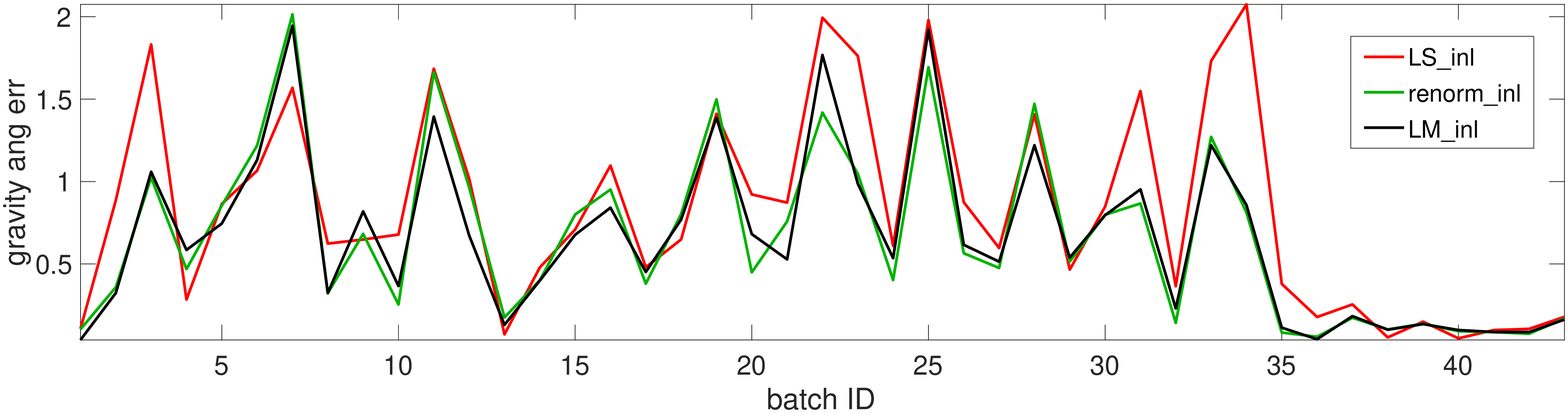} \\
      \footnotesize{(d)} \\\vspace*{-5mm}
    \end{center}
 \caption{\MM{OfficeLoop} sequence. Left stereo image (a)~at the start and (b)~in the middle of the sequence. Trajectory is the same as in \Fig{fig:result_trapcam}(c). See \Fig{fig:result_subway} for the remaining caption. }
 \label{fig:result_office_loop}
\end{figure}

We report quantitative results in \Tab{tab:results} for the six sequences, shown in \Fig{fig:result_subway}, \Fig{fig:result_trapcam}, and \Fig{fig:snapshots_rendered_seqs}. The "forward" trajectory is shown in \Fig{fig:result_subway}, the "loop" trajectory in \Fig{fig:result_trapcam}. The "fast shaking" trajectory is $0.5$\,m wide left-right shaking motion with rapid acceleration and average velocity of $0.9\,\textrm{ms}^{-1}$. The "for/back-ward" trajectory is $14$\,m straight forward, followed by $180^\circ$ turn and back to the start with the average speed $1.8\,\textrm{ms}^{-1}$. We captured these typical motions of a person when wearing smart glasses when moving in the office space shown in \Fig{fig:result_office_loop}.

We present detailed qualitative results for two sequences. The first sequence, \MM{SubwayTrain} is a forward  $9$\,m long sequence inside a static subway train, see \Fig{fig:result_subway}.  The second sequence,  \MM{TrapCam} is a loop shaped $25$\,m long sequence outdoors, see \Fig{fig:result_trapcam}. 


The results confirm the observation from the Synthetic experiment that  \MM{ls} method can be improved by the renormalization \MM{rnm} which is comparable and sometimes better to ML estimation of \MM{ba}. In most cases, the initial velocity $\V{v}_0$ and gravity $\V{g}_0$ are both improved \wrt the \MM{ls}, and this by roughly $20\%$ and $8\%$, respectively. This is a significant improvement.

\subsection{Real Data}


\begin{figure}[t]
  \begin{center}
    \begin{tabular}{@{\hspace{0mm}}c@{\hspace{2mm}}c}
       \includegraphics[width=0.49\linewidth]{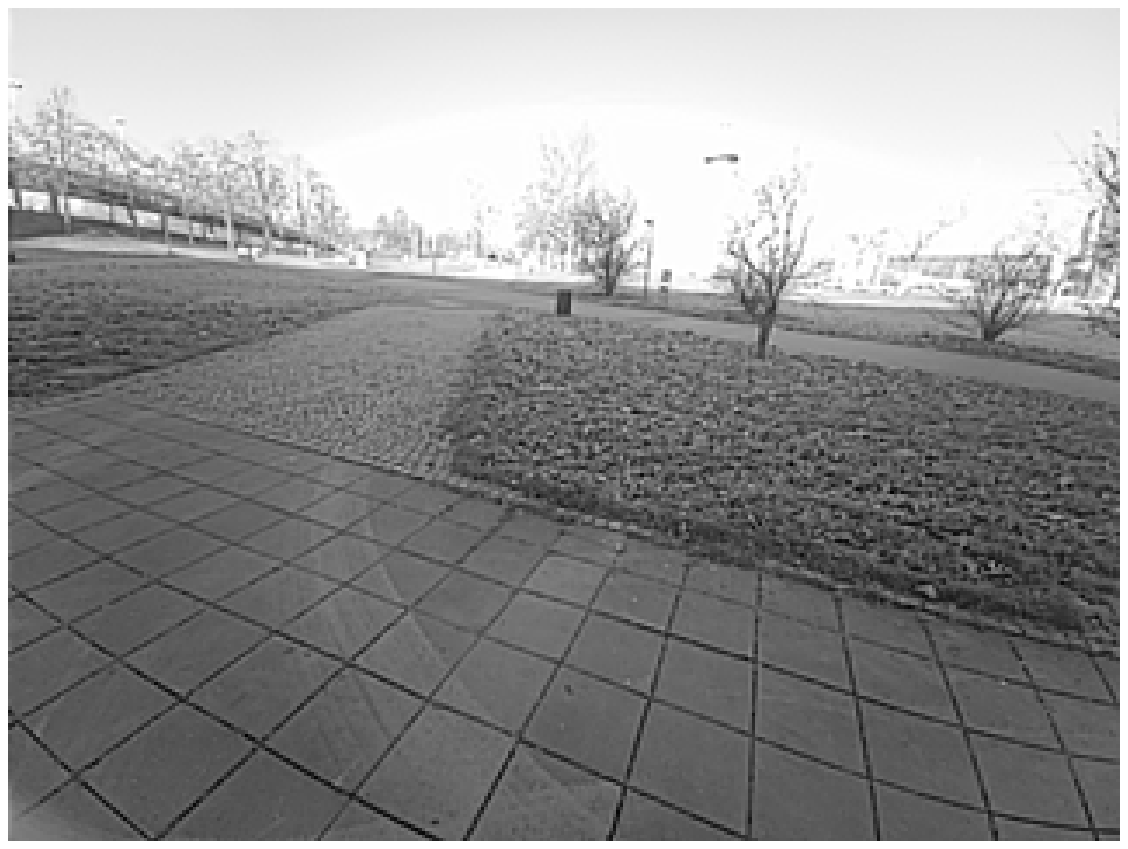} &
       \includegraphics[width=0.49\linewidth]{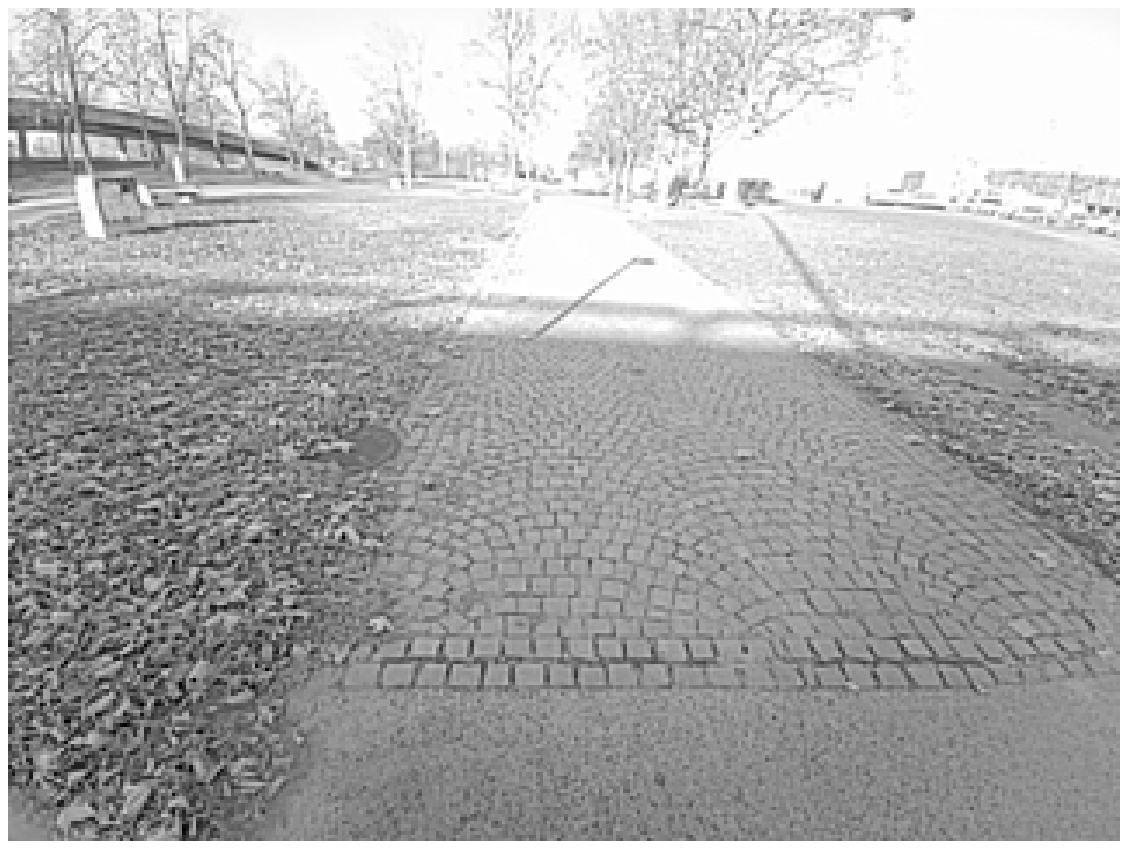} \\
       (a) & (b)\\[1ex]
     \end{tabular} 
     \scriptsize
     \psfrag{x [m]}[bc][bc]{$x [\textrm{m}]$}
     \psfrag{y [m]}[bc][bc]{$y [\textrm{m}]$}
     \includegraphics[width=0.98\linewidth, trim = 1 0 0 1, clip=true]{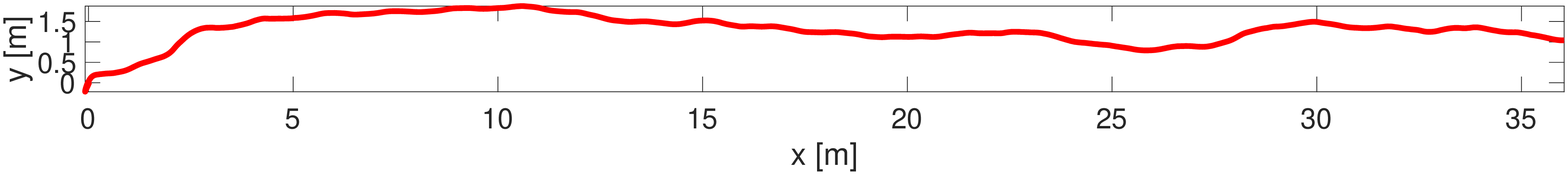} \\[2ex]
     \psfrag{velocity [m/s]}[bc][bc]{velocity [m/s]}
     \includegraphics[width=0.98\linewidth, trim = 1 0 0 1, clip=true]{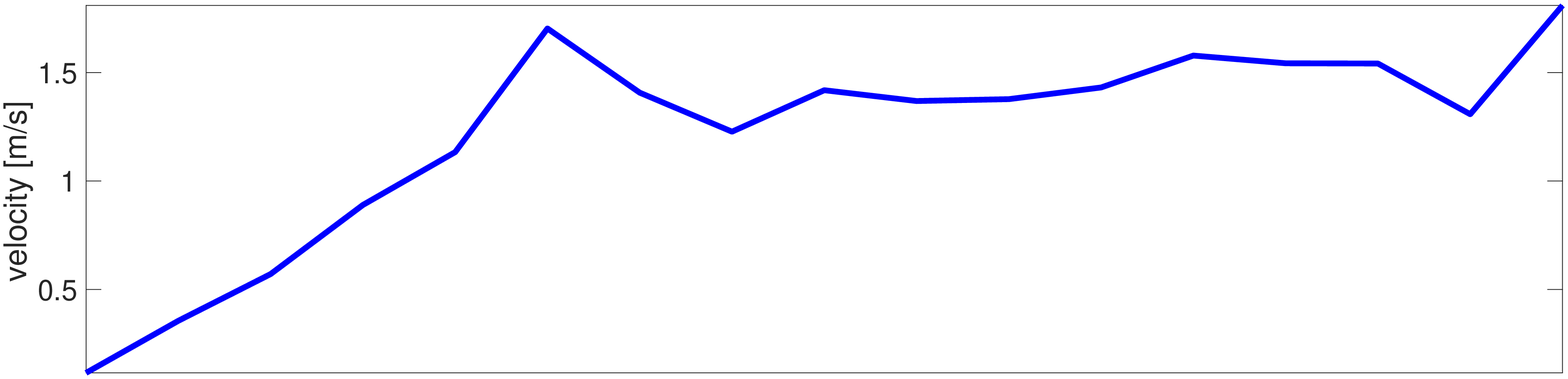}\\
     (c)\\[2ex]
    \scriptsize
    \psfrag{LS_inl}[l][l]{\MM{ls}}
    \psfrag{renorm_inl}[l][l]{\MM{rnm}}
    \psfrag{LM_inl}[l][l]{\MM{ba}}
    \psfrag{velocity dst err}[bc][bc]{$\epsilon_{\V{v}_0}$ $[\textrm{m/s}]$}
    \psfrag{batch ID}[tc][tc]{batch ID}
    \includegraphics[width=0.98\linewidth,  trim = 1 0 0 1, clip=true]{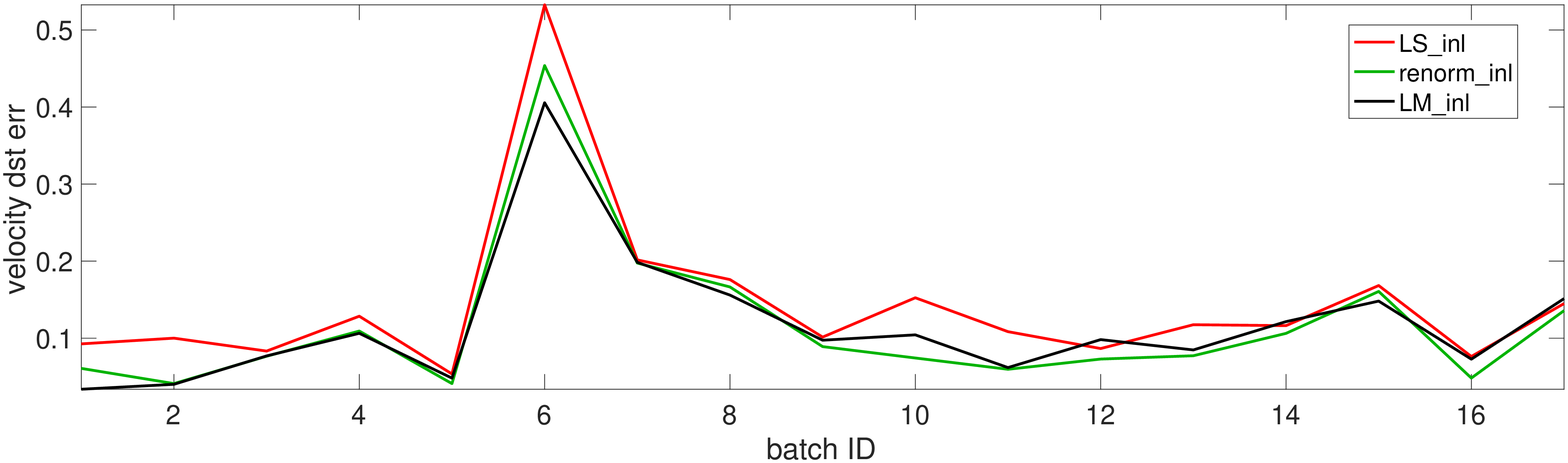} \\[2ex]
    \psfrag{gravity ang err}[bc][bc]{$\epsilon_{\V{g}_0}$[deg]}
     ~\includegraphics[width=0.98\linewidth, trim = 1 0 0 1, clip=true ]{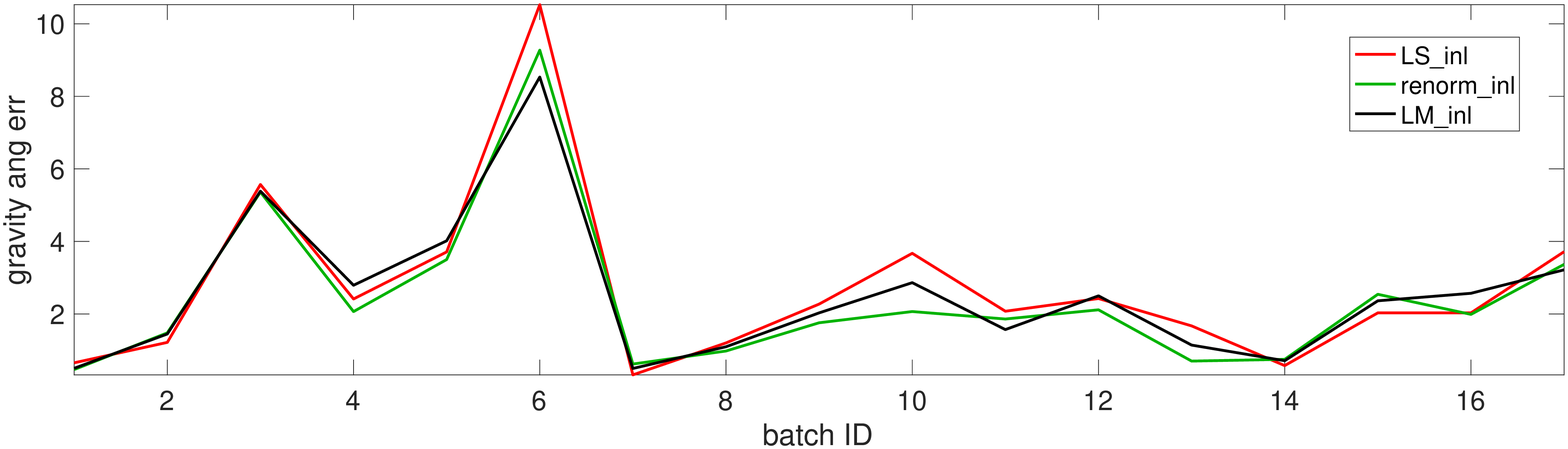} \\
     \footnotesize{(d)} \\\vspace*{-5mm}
    \end{center}
 \caption{\MM{Walk} sequence. Left stereo image (a)~at the start and (b)~in the middle of the sequence. See \Fig{fig:result_subway} for the remaining caption. }
 \label{fig:result_walk12}
\end{figure}

We use real data from the Snap Spectacles glasses, as stereo images as well as \imu{} readings. We do not posses
ground truth for these sequences. Instead, we run a typical VIO system based on the temporal Extended Kalman Filter, similar to \cite{Mourikis-ICRA2007, Li-ICRA2013}. The filter framework fuses inertial and visual data in iterative updating procedure for maximum a posteriori probability of a linear dynamical system. The filter uses a strong prior that the sequences are static at the beginning, copes with a rolling shutter stereo camera and optimizes also for both accelerometer and gyroscope biases. For the proposed solver, though, we do not include the biases as we found that their magnitude is low in the used device.

The first is the \MM{OfficeLoop}, a loop-shaped $25$\,m long sequence in a typical open space office. Since there is only negligible drift between the end and starting position, we can consider the used \MM{vio} as a reasonably accurate baseline to compare to. The second \MM{Walk} is a forward $36$\,m long sequence outdoors. Both sequences are acquired during a walk.

The results align with the previous synthetic and rendered experiments; the renormalization \MM{rnm} is similar to ML estimation of \MM{ba}, both outperforming the Least Squares \MM{ls}. As mentioned, we do not have the ground truth and the comparison for these sequences might not be representative. What should be noticed and taken from these results is that the renormalization and ML estimator perform very similarly to each other, although, arriving to the solution by different means. 

\subsection{Complexity}

The renormalization scheme and Bundle Adjustment require different operation flow which yields different complexity. We give hints to the expected performance by pinpointing the most time consuming parts during the computations. 

{\bf Renormalization.} Complexity of renormalization is driven by computation of the partial derivatives in \Eq{eq:Gderivative} which is needed for the covariance matrix in \Eq{eq:cov_propagation} to fill $\M{M}$ and $\M{N}$ matrices. The involved matrices in \Eq{eq:Gderivative} are sparse with derivatives of $\M{P}$ consisting of a few ones ($4$ entries for a point tracked in $5$ frames). Taking this into account yields many savings in computation. The generalized eigenvalue problem on $7 \times 7$ matrix itself is negligible, and typically only 3 iterations suffices.

{\bf Bundle Adjustment.} Complexity in Levenberg-Marquardt optimizer is spread roughly equally into three parts. First, building a sparse linear system of normal equations with $2N \times (3M+6)$ matrix and its corresponding right hand side vector, where $N$ is the number of observations and $M$ the number of auxiliary 3D points. Second, making the matrix square by left multiplying with its transpose. Third, running a sparse linear solver. Typically, $15$ iterations are needed.

With our \MM{Matlab} implementations, the tests on the presented rendered and real sequences show that the \MM{rnm} method takes on average around $70$\% of the time of \MM{ba}, depending on the number of considered frame pairs and length of the tracks. Our \texttt{C++} implementation of the baseline \MM{ls} method takes on average $\sim6$ms and of \MM{ba} $\sim25$ms on an i7@2.6GHz CPU, given input inlier observations, considering five stereo frames with their pairing detailed in \Sec{sec:syn_experiments}.

\section{Conclusion}
\label{sec:conclusion}

We presented a novel way to solve the initialization problem of the inertial-visual odometry system. We derived a novel solver through proper statistical modeling and we cast the problem into the renormalization scheme of Kanatani. We incorporated proper noise propagation thus yielding a solution which exhibits higher accuracy over the original Least Squares solution. The extensive evaluation shows that the renormalization scheme performs very closely to the ML estimator which is statistically optimal in case of Gaussian noise. As such, the renormalization can serve to get a very good initial point for the ML, or fully replace it, as the additional improvement is rather small for the cost of more computations. 

With this paper, we add a new problem into the set of problems in Computer Vision which can be beneficially solved by the renormalization scheme. As the set of problems where the renormalization improves Gold Standards grows, the renormalization scheme is slowly finding its way into the Computer Vision community.
\bibliographystyle{spbasic}
\bibliography{vioInitRenorm}

\begin{thebibliography}{42}
\providecommand{\natexlab}[1]{#1}
\providecommand{\url}[1]{{#1}}
\providecommand{\urlprefix}{URL }
\expandafter\ifx\csname urlstyle\endcsname\relax
  \providecommand{\doi}[1]{DOI~\discretionary{}{}{}#1}\else
  \providecommand{\doi}{DOI~\discretionary{}{}{}\begingroup
  \urlstyle{rm}\Url}\fi
\providecommand{\eprint}[2][]{\url{#2}}

\bibitem[{Albl et~al(2015)Albl, Kukelova, and Pajdla}]{Albl-CVPR2015}
Albl C, Kukelova Z, Pajdla T (2015) {R6P - Rolling Shutter Absolute Pose
  Problem}. In: Proc. CVPR

\bibitem[{{Albl} et~al(2016){Albl}, {Kukelova}, and {Pajdla}}]{Albl-CVPR2016}
{Albl} C, {Kukelova} Z, {Pajdla} T (2016) Rolling shutter absolute pose problem
  with known vertical direction. In: Proc. CVPR

\bibitem[{{Apple}(2015)}]{ARKit}
{Apple} (2015) {ARKit}. \url{https://developer.apple.com/augmented-reality}

\bibitem[{Bapat et~al(2018)Bapat, Price, and Frahm}]{Bapat-CVPR2018}
Bapat A, Price T, Frahm JM (2018) Rolling shutter and radial distortion are
  features for high frame rate multi-camera tracking. In: Proc. CVPR

\bibitem[{Campos et~al(2019)Campos, Montiel, and Tard{\'o}s}]{Campos-ICRA2019}
Campos C, Montiel J, Tard{\'o}s J (2019) {Fast and Robust Initialization for
  Visual-Inertial SLAM}. In: Proc. ICRA

\bibitem[{Chojnacki et~al(2000)Chojnacki, Brooks, van~den Hengel, and
  Gawley}]{Chojnacki-PAMI2000}
Chojnacki W, Brooks MJ, van~den Hengel A, Gawley D (2000) On the fitting of
  surfaces to data with covariances. PAMI 22(11)

\bibitem[{Chojnacki et~al(2001)Chojnacki, Brooks, and van~den
  Hengel}]{Chojnacki-JMIV2001}
Chojnacki W, Brooks MJ, van~den Hengel A (2001) Rationalising the
  renormalisation method of {K}anatani. {Journal of Mathematical Imaging and
  Vision} 14:21--38

\bibitem[{Dai et~al(2016)Dai, Li, and Kneip}]{Dai-CVPR2016}
Dai Y, Li H, Kneip L (2016) Rolling shutter camera relative pose: Generalized
  epipolar geometry. In: Proc. CVPR

\bibitem[{{Dong-Si} and {Mourikis}(2012)}]{Dong-IROS2012}
{Dong-Si} T, {Mourikis} AI (2012) Estimator initialization in vision-aided
  inertial navigation with unknown camera-imu calibration. In: IEEE/RSJ
  International Conference on Intelligent Robots and Systems

\bibitem[{{Epic Games}(2019)}]{Unreal}
{Epic Games} (2019) {Unreal Engine}. \url{http://www.unrealengine.com}

\bibitem[{Evangelidis and Micusik(2021)}]{Evangelidis-RAL2021}
Evangelidis G, Micusik B (2021) Revisiting visual-inertial
  structure-from-motion for odometry and {SLAM} initialization. Robotics and
  Automation Letters (RA-L) 6(2)

\bibitem[{{Evangelidis} and {Psarakis}(2008)}]{Evangelidis-PAMI2008}
{Evangelidis} GD, {Psarakis} EZ (2008) Parametric image alignment using
  enhanced correlation coefficient maximization. PAMI 30(10)

\bibitem[{Forster et~al(2017)Forster, Carlone, Dellaert, and
  Scaramuzza}]{Forster-TRO2017}
Forster C, Carlone L, Dellaert F, Scaramuzza D (2017) On-manifold
  preintegration for real-time visual--inertial odometry. IEEE TRO 33(1)

\bibitem[{F{\"o}rstner and Wrobel(2016)}]{Foerstner-Book2016}
F{\"o}rstner W, Wrobel B (2016) Photogrammetric Computer Vision. Springer

\bibitem[{Golub and van Loan(2013)}]{Golub-Matrix2013}
Golub GH, van Loan CF (2013) Matrix Computations, 4th edn. JHU Press

\bibitem[{{Google}(2018)}]{ARCore}
{Google} (2018) {ARCore}. \url{https://developers.google.com/ar}

\bibitem[{Gupta and Hartley(1997)}]{Gupta-PAMI1997}
Gupta R, Hartley RI (1997) Linear pushbroom cameras. PAMI 19(9)

\bibitem[{Hartley and Zisserman(2004)}]{Hartley-Book2004}
Hartley RI, Zisserman A (2004) Multiple View Geometry in Computer Vision.
  Cambridge University Press

\bibitem[{Hedborg et~al(2012)Hedborg, Forssen, Felsberg, and
  Ringaby}]{Hedborg-CVPR2012}
Hedborg J, Forssen PE, Felsberg M, Ringaby E (2012) Rolling shutter bundle
  adjustment. In: Proc. CVPR

\bibitem[{{Huang} et~al(2020){Huang}, {Liu}, and {Wan}}]{Huang-TRO2020}
{Huang} W, {Liu} H, {Wan} W (2020) An online initialization and
  self-calibration method for stereo visual-inertial odometry. TRO Preprint

\bibitem[{{Kaiser} et~al(2017){Kaiser}, {Martinelli}, {Fontana}, and
  {Scaramuzza}}]{Kaiser-RAL2017}
{Kaiser} J, {Martinelli} A, {Fontana} F, {Scaramuzza} D (2017) Simultaneous
  state initialization and gyroscope bias calibration in visual inertial aided
  navigation. IEEE Robotics and Automation Letters 2(1):18--25

\bibitem[{Kanatani(1996)}]{Kanatani-Stat1996}
Kanatani K (1996) Statistical Optimization for Geometric Computation: Theory
  and Practice. Elsevier Science Inc., USA

\bibitem[{Kanatani(2008)}]{Kanatani-IJCV2008}
Kanatani K (2008) Statistical optimization for geometric fitting: Theoretical
  accuracy bound and high order error analysis. IJCV 80

\bibitem[{{Kanatani}(2014)}]{Kanatani-ICPR2014}
{Kanatani} K (2014) Statistical optimization for geometric estimation:
  Minimization vs. non-minimization. In: Proc. ICPR

\bibitem[{Kanatani et~al(2016)Kanatani, Sugaya, and
  Kanazawa}]{Kanatani-Guide3D}
Kanatani K, Sugaya Y, Kanazawa Y (2016) Guide to 3D Vision Computation.
  {Springer Verlag}

\bibitem[{{Kneip} et~al(2011){Kneip}, {Weiss}, and {Siegwart}}]{Kneip-IROS2011}
{Kneip} L, {Weiss} S, {Siegwart} R (2011) Deterministic initialization of
  metric state estimation filters for loosely-coupled monocular vision-inertial
  systems. In: IEEE/RSJ International Conference on Intelligent Robots and
  Systems

\bibitem[{Leedan and Meer(2000)}]{Leedan-IJCV2000}
Leedan Y, Meer P (2000) {Heteroscedastic Regression in Computer Vision:
  Problems with Bilinear Constraint}. IJCV 37(2)

\bibitem[{Li et~al(2013)Li, Kim, and Mourikis}]{Li-ICRA2013}
Li M, Kim B, Mourikis A (2013) Real-time motion tracking on a cellphone using
  inertial sensing and a rolling-shutter camera

\bibitem[{{Ling} et~al(2018){Ling}, {Bao}, {Jie}, {Zhu}, {Li}, {Tang}, {Liu},
  {Liu}, and {Zhang}}]{Ling-ECCV2018}
{Ling} Y, {Bao} L, {Jie} Z, {Zhu} F, {Li} Z, {Tang} S, {Liu} Y, {Liu} W,
  {Zhang} T (2018) Modeling varying camera-imu time offset in
  optimization-based visual-inertial odometry. In: Proc. ECCV

\bibitem[{{Lourakis} and {Argyros}(2005)}]{Loukaris-ICCV2005}
{Lourakis} MLA, {Argyros} AA (2005) {Is Levenberg-Marquardt the most efficient
  optimization algorithm for implementing bundle adjustment?} In: Proc. ICCV,
  vol~2

\bibitem[{Martinelli(2013)}]{Martinelli-IJCV2013}
Martinelli A (2013) {Closed-form solution of visual-inertial structure from
  motion}. IJCV

\bibitem[{Meingast et~al(2005)Meingast, Geyer, and Sastry}]{Meingast-CoRR2005}
Meingast M, Geyer C, Sastry S (2005) Geometric models of rolling-shutter
  cameras. CoRR

\bibitem[{{Mourikis} and {Roumeliotis}(2007)}]{Mourikis-ICRA2007}
{Mourikis} AI, {Roumeliotis} SI (2007) A multi-state constraint kalman filter
  for vision-aided inertial navigation. In: Proc. ICRA

\bibitem[{{Mur-Artal} and {Tard{\'o}s}(2017)}]{Mur-Artal-RAL2017}
{Mur-Artal} R, {Tard{\'o}s} JD (2017) Visual-inertial monocular slam with map
  reuse. IEEE Robotics and Automation Letters 2(2)

\bibitem[{{Mur-Artal} et~al(2015){Mur-Artal}, {Montiel}, and
  {Tard{\'o}s}}]{Mur-Artal-TRO2015}
{Mur-Artal} R, {Montiel} J, {Tard{\'o}s} JD (2015) {ORB-SLAM: a versatile and
  accurate monocular slam system}. TRO 31(5)

\bibitem[{{Okatani} and {Deguchi}(2009)}]{Okatani-CVPR2009}
{Okatani} T, {Deguchi} K (2009) On bias correction for geometric parameter
  estimation in computer vision. In: Proc. CVPR

\bibitem[{Patron-Perez et~al(2015)Patron-Perez, Lovegrove, and
  Sibley}]{Patron-Perez-IJCV2015}
Patron-Perez A, Lovegrove S, Sibley G (2015) A spline-based trajectory
  representation for sensor fusion and rolling shutter cameras. IJCV 113

\bibitem[{{Qin} and {Shen}(2017)}]{Qin-IROS2011}
{Qin} T, {Shen} S (2017) Robust initialization of monocular visual-inertial
  estimation on aerial robots. In: IEEE/RSJ International Conference on
  Intelligent Robots and Systems

\bibitem[{Rosten et~al(2010)Rosten, Porter, and Drummond}]{Rosten-PAMI2008}
Rosten E, Porter R, Drummond T (2010) {FASTER and better: A machine learning
  approach to corner detection}. PAMI 32:105--119

\bibitem[{Schubert et~al(2018)Schubert, Demmel, Usenko, St{\"u}ckler, and
  Cremers}]{Schubert-ECCV2018}
Schubert D, Demmel N, Usenko V, St{\"u}ckler J, Cremers D (2018) Direct sparse
  odometry with rolling shutter. In: Proc. ECCV

\bibitem[{Schubert et~al(2019)Schubert, Demmel, Stumberg, Usenko, and
  Cremers}]{Schubert-IROS2019}
Schubert D, Demmel N, Stumberg L, Usenko V, Cremers D (2019) Rolling-shutter
  modelling for direct visual-inertial odometry

\bibitem[{{Taubin}(1991)}]{Taubin-PAMI1991}
{Taubin} G (1991) Estimation of planar curves, surfaces, and nonplanar space
  curves defined by implicit equations with applications to edge and range
  image segmentation. PAMI 13(11)

\end{thebibliography}


\end{document}